\documentclass[lettersize,journal]{IEEEtran}
\usepackage{packages}

\title{A Hierarchical, Model-Based System for High-Performance Humanoid Soccer}

\author{Quanyou Wang$^{1\dagger}$, Mingzhang Zhu$^{1\dagger}$, Ruochen Hou$^{1\dagger}$, Kay Gillespie$^{1}$, Alvin Zhu$^{1}$, Shiqi Wang$^{1}$, Yicheng Wang$^{1}$, Gaberiel I. Fernandez$^{1,2}$, Yeting Liu$^{1}$, Colin Togashi$^{1,2}$, Hyunwoo Nam$^{1}$, Aditya Navghare$^{1,2}$, Alex Xu$^{1}$, Taoyuanmin Zhu$^{1}$, Min Sung Ahn$^{1}$, Arturo Flores Alvarez$^{1}$, Justin Quan$^{1}$, Ethan Hong, and Dennis W. Hong$^{1}$

    \thanks{$^{1}$Department of Mechanical and Aerospace Engineering, UCLA, Los Angeles, CA, USA.}
    \thanks{$^{2}$VEQDrive, the Robotics and AI Startup.}
    \thanks{$^{\dagger}$ Equal contribution.}
}

\begin{document}
\maketitle

\begin{abstract}

The development of athletic humanoid robots has gained significant attention as advances in actuation, sensing, and control enable increasingly dynamic, real-world capabilities. RoboCup, an international competition of fully autonomous humanoid robots, provides a uniquely challenging benchmark for such systems, culminating in the long-term goal of competing against human soccer players by 2050. This paper presents the hardware and software innovations underlying our team’s victory in the RoboCup 2024 Adult-Sized Humanoid Soccer Competition. On the hardware side, we introduce an adult-sized humanoid platform built with lightweight structural components, high-torque quasi-direct-drive actuators, and a specialized foot design that enables powerful in-gait kicks while preserving locomotion robustness. On the software side, we develop an integrated perception and localization framework that combines stereo vision, object detection, and landmark-based fusion to provide reliable estimates of the ball, goals, teammates, and opponents. A mid-level navigation stack then generates collision-aware, dynamically feasible trajectories, while a centralized behavior manager coordinates high-level decision making, role selection, and kick execution based on the evolving game state. The seamless integration of these subsystems results in fast, precise, and tactically effective gameplay, enabling robust performance under the dynamic and adversarial conditions of real matches. This paper presents the design principles, system architecture, and experimental results that contributed to ARTEMIS’s success as the 2024 Adult-Sized Humanoid Soccer champion.

\end{abstract}


\section{Introduction}
\label{sec:introduction}
Humanoid robotics has made rapid progress in recent years, driven by advances in both hardware and control. The emergence of quasi-direct-drive (QDD) actuators and torque-controlled joints have enabled robots to exhibit fast-response, dynamic, impact-resilient behaviors while generating substantial torque \cite{2019_tym_liquid_cooled_motor, MIT_Cheetah_actuator, 2024_adrian_icra, 2024_alvin_icra}. At the same time, improved whole-body locomotion controllers, optimal control methods, and learning-based techniques have expanded the range of dynamic motions that humanoids can execute, from robust bipedal walking to running and acrobatic maneuvers \cite{ahn2023_thesis, lijunheng_FMMPC, huang2024diffuselocomotion, 2025_yusuke_humanoids, 2025_qiayuan_icra}. These developments have enabled humanoid robots to tackle an increasingly diverse set of tasks, such as loco-manipulation, assistance in human environments, and life-like performance and entertainment \cite{li2023locomanipulationhector, 2019_HRP_5P, 2025kidcosmo_main}. While many of these applications occur in relatively structured settings or involve short-horizon, partly scripted behaviors, humanoid robot soccer stands out as an especially demanding testbed: it requires perception, locomotion, planning, and decision making to be integrated into a single, long-horizon, fully autonomous task.

\begin{figure*} [t!]
    \centering
    \includegraphics[width=\linewidth]{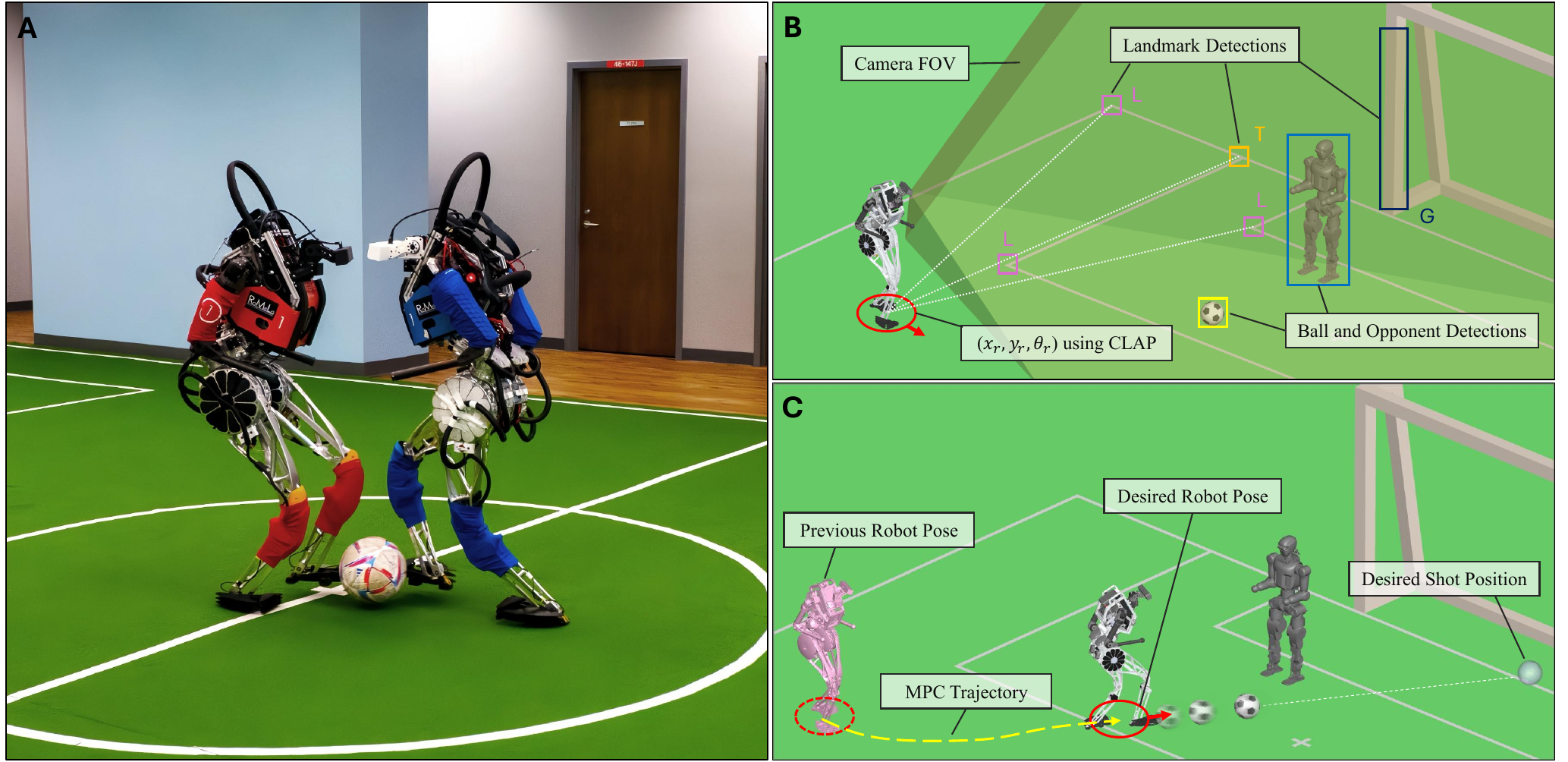}
    \caption{Overview of the ARTEMIS humanoid soccer system. A). Two ARTEMIS humanoid robots competing for ball possession during a practice match. B). Perception and localization pipeline: stereo vision detects field landmarks, the ball, teammates, and opponents; CLAP fuses landmark geometry with inertial measurements to estimate the robot pose ($x_{r}$, $y_{r}$, $\theta_{r}$). Proximity sensing provides complementary obstacle information when vision is degraded. C). Mid-level navigation and behavior framework: the robot selects a desired pose and shot target based on the game state, and a collision-aware MPC controller generates a dynamically feasible trajectory that respects proximity constraints and field geometry.}
    \label{fig:cover}
\end{figure*}

Despite significant advancements in humanoid robotics, competitive soccer presents a uniquely demanding benchmark that stresses every layer of a robot’s design. A humanoid must generate coherent, long-horizon behaviors while continually adapting to an environment that changes unpredictably. Visual perception is noisy, delayed, and frequently disrupted by motion, occlusions, and lighting variations, while the controller must remain stable under falls, collisions, and rapid changes in ball possession. Effective performance therefore requires tight integration of perception, localization, footstep planning, and whole-body control to maintain stability, track dynamic targets, and execute precise kicks.

These challenges are further amplified in adult-sized humanoid soccer. Larger robots are heavier, taller, and capable of producing substantial forces, making balance recovery, impact mitigation, and rapid momentum redirection significantly more demanding. The RoboCup Humanoid League also restricts robots to onboard cameras and proprioception, forcing all perception, planning, and decision making to run autonomously under limited sensing \cite{Kitano1997RoboCup}. Combined with soft turf, fast-moving opponents, and a compact field that encourages frequent physical interactions, these constraints make adult-sized humanoid soccer one of the most demanding real-world settings for integrated perception, planning, and dynamic locomotion \cite{2018nimbro_ros_framework, Forero2014ROSNAO}.

Many RoboCup teams have developed adult-sized humanoids using servo-based actuation and ZMP-style locomotion. These systems enable statically stable, medium-speed walking and basic kicking, but they struggle with fast, dynamic motions, limiting both mobility on the field and scoring efficiency. Although strong in-gait kicks are among the most effective ways to score, they are difficult to execute because the required leg accelerations and sudden impact forces can easily destabilize the robot. As a result, most robots still rely on semi-static kicks or weak in-gait kicks, which restricts both scoring performance and tactical flexibility \cite{robocup2023_nimbro_championship, 2021_6D_LocalizationKicking, Mueller2010Kicking}.

From a software perspective, existing RoboCup humanoid stacks exhibit notable limitations. Many competitive architectures, use simple behavior planners with limited role reasoning and lack a mid-level navigation layer for online global path planning \cite{robocup2023_nimbro_championship}. Proximity sensing and collision-avoidance capabilities are often minimal or absent, leaving robots vulnerable to interference and congestion during close interactions. In addition, vision, neck control, and localization are typically treated as loosely coupled modules rather than coordinating both objectives through unified sensing strategies. Such weak integration between perception and control, also observed in earlier modular robot architectures \cite{2006_robot_system}, makes it difficult to maintain consistent, high-performance behavior under the dynamic and adversarial conditions of real matches.

Recent learning-based approaches have explored tighter perception–action coupling for humanoid soccer. Deep RL policies have been trained to map compact visual observations—typically limited to ball and goal features—directly to joint commands for reactive ball pursuit and shooting \cite{booster_2025learning, haarnoja_learning_2024}. These methods demonstrate impressive single-robot performance but focus on isolated skills without modeling teammates, opponents, or obstacles. Consequently, they must be embedded within larger, hand-engineered stacks to handle coordination, strategy, and collision-aware navigation. Overall, these RL skill modules, like classical modular stacks, do not yet provide a fully integrated, field-ready software architecture for adult-sized humanoid soccer.

In this work, we present ARTEMIS (Advanced Robotic Technology for
Enhanced Mobility and Improved Stability), a fully integrated adult-sized humanoid soccer platform designed to address these challenges. On the hardware side, specialized foot attachments combined with high-torque, high-acceleration QDD actuators and in-gait locomotion control enable powerful kicks while walking, eliminating the need to stop and reconfigure the stance. On the perception side, our integrated vision–proximity–localization framework provides robust and accurate estimates of the ball, goals, teammates, and opponents. Proximity sensing serves as a complementary modality, offering redundancy when vision degrades and supplying inputs for collision avoidance. Building on these perception and localization capabilities, we introduce a mid-level planner that generates collision-free, dynamically feasible trajectories in real time using proximity data and field geometry. At the highest level, we design a behavior architecture that tightly couples desired-state planning, memory handling, motion prediction, and specialized managers for kicking and neck control. This controller coordinates actions such as repositioning, turning, ball approach, in-gait kicking, and recovery, allowing ARTEMIS to transition smoothly between game states and adapt rapidly to environmental changes. Finally, we evaluate the complete system in RoboCup 2024 and additional controlled experiments, demonstrating through quantitative and qualitative results that the tight integration of these components enables robust, agile, and tactically effective performance at a level sufficient to win an international adult-sized humanoid soccer competition.

\section{System Overview}
\label{sec:System Overview}

Our framework is designed as a tightly integrated perception–planning–control stack for dynamic humanoid soccer. \cref{fig:cover} illustrates how these capabilities appear on the field: the robot perceives landmarks, the ball, and opponents, localizes itself on the pitch, and executes collision-aware motions to a desired pose and shot target. Building on our RoboCup 2024 champions guide \cite{2024_robocup_symposium} and the open-source ARTEMIS platform \cite{artemis_open_source}, this section details the system architecture that enables such behavior.

\begin{figure*}[t!]
    \centering
    \includegraphics[width=\linewidth]{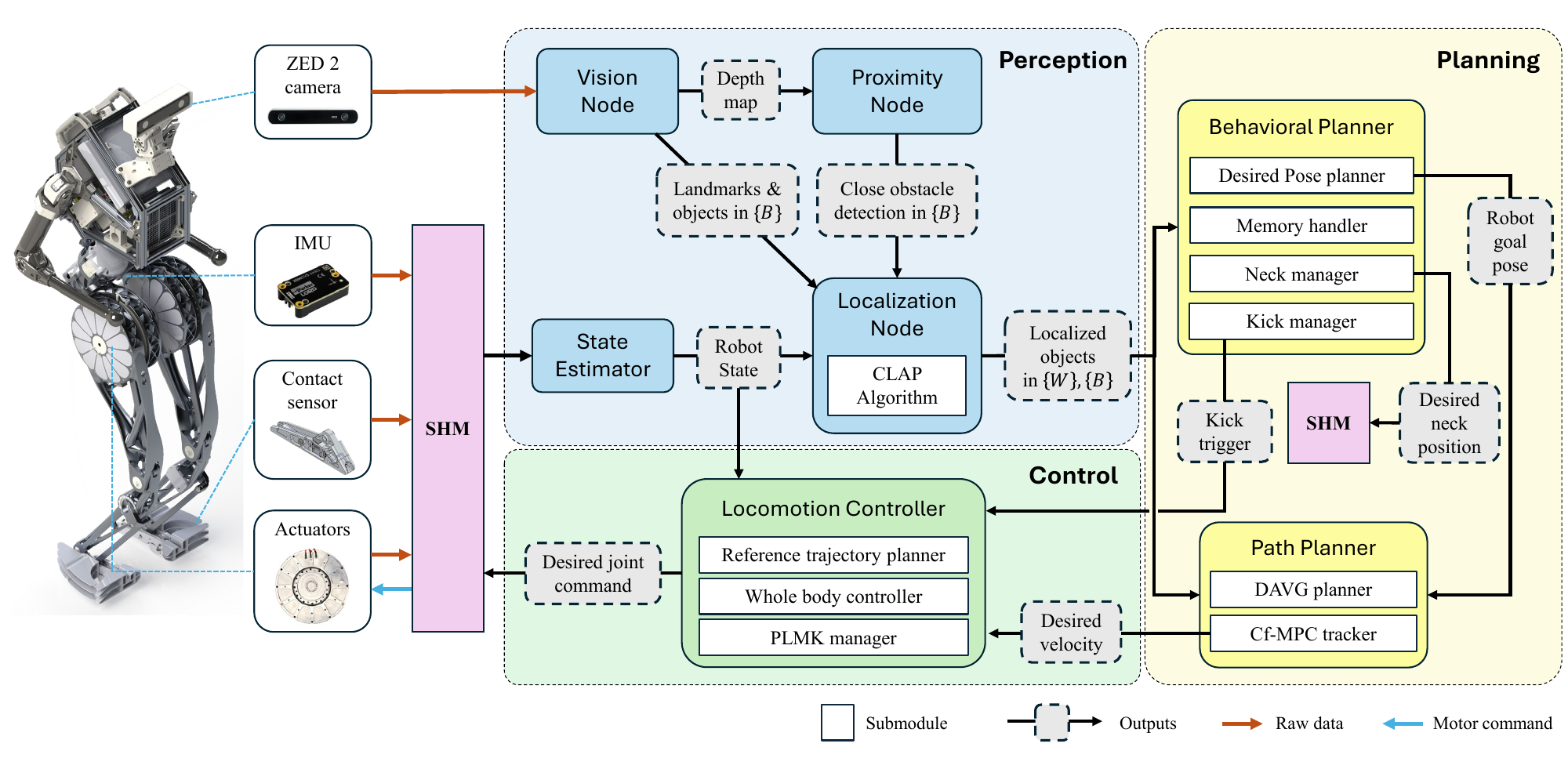}
    \caption{System architecture of the ARTEMIS humanoid platform. The perception layer provides object detections, proximity cues, and pose estimates that feed into the path planner and behavioral planner module. A high-rate shared-memory loop (SHM) handles hardware interaction and executes the locomotion controller. }
    \label{fig:Control_framework}
\end{figure*}

The software framework itself is summarized in \cref{fig:Control_framework}. The architecture is organized into a fast, shared-memory control loop and a set of ROS~2 nodes for perception, localization, mid-level navigation, and high-level behavior planning. Proprioceptive data from the actuators, IMU, and foot contact sensors are fused in a high-rate estimator, which provides contact and robot state to the locomotion controller at 1~kHz. This low-level layer executes whole-body control and in-gait kicking on the ARTEMIS hardware \cite{artemis_open_source}, ensuring that balance and contact transitions remain stable even when higher-level modules update at slower rates.

On top of this real-time loop, the perception and localization stack operates as ROS~2 nodes. A stereo vision node processes images from the ZED~2 camera at about 60~Hz to generate depth maps and detections of field landmarks, the ball, teammates, and opponents. A proximity node extracts close-range obstacle information from depth data, providing a robust signal for nearby robots and serving as a backup when visual detections are partially occluded. These observations are expressed in the robot frame and passed to the localization node. The localization module implements CLAP-based geometric localization \cite{2025_hou_clap_localization, hou2025_fastrobust_localization} for humanoid soccer, producing globally consistent estimates of robot and object states in both world and robot frames. This combination yields pose estimates that remain reliable even under the symmetric field layout and frequent occlusions characteristic of RoboCup matches.

Using these state estimates, the mid-level navigation and behavior layer plans and executes game-level actions. The midlevel node couples a \emph{Dynamic Augmented Visibility Graphs} (DAVG) planner with a \emph{Collision-free Model Predictive Controller} (cf-MPC) tracker, following \cite{hou2025icra}, to produce dynamically feasible trajectories that account for field boundaries and moving obstacles. At the top of the stack, a centralized behavior planner node selects tactical actions and coordinates the interaction between perception, navigation, and locomotion. The behavior planner maintains an internal memory of ball and opponent states, predicts short-horizon motion, and selects desired robot poses, roles, and kicking actions based on the evolving game context. Specialized managers for kicking and neck control convert these decisions into concrete commands: the kick manager chooses kick type and timing while the neck manager directs gaze to balance ball tracking and landmark observations. Through shared memory interfaces, these high-level decisions are synchronized with the mid-level planner and low-level controller, ensuring smooth transitions between behaviors such as repositioning, turning, ball approach, and in-gait kicking.

\section{Hardware}
\label{sec:hardware}

\subsection{Hardware Platform}
ARTEMIS, which features a high-performance humanoid design with 5 degrees of freedom (DoF) per leg, 4 DoF per arm, and 2 DoF in the neck, is the platform we used during RoboCup. The robot is equipped with proprioceptive actuators that exhibit low reflected inertia and high transmission transparency. These characteristics are essential for impact mitigation and precise force control, enabling robust and dynamic locomotion in competitive scenarios.

The 5-DoF leg design intentionally omits the ankle roll joint to reduce distal leg mass and enhance leg acceleration during the swing phase, improving performance in rapid walking, running, and disturbance recovery. While this limits single-leg balancing, such postures are not required for our applications and this simplification benefits model-based control. To equalize torque requirements while increasing hip pitch range for human-like squatting, the hip roll and yaw axes are tilted forward by 45°. Distal mass is further reduced by placing actuators near the torso through parallel four-bar linkages, avoiding the compliance and inconsistency associated with belts and chains. As a result, over 60\% of the leg mass is concentrated near the hip joint, significantly improving acceleration and stability. Furthermore, actuator housings are integrated directly into the load-bearing structure with topology-optimized femur and tibia components, minimizing fasteners and excess material while maintaining high strength-to-weight ratios.



For autonomous soccer operation, ARTEMIS is equipped with a ZED 2 stereo camera for environmental perception and a MicroStrain 3DM-CV7 IMU located in the lower torso for balance estimation and motion stabilization. Computational tasks are executed on an HP Elite Mini 800 G9 computer mounted on the robot, equipped with an Intel Core i7 processor and an NVIDIA RTX 3050 Ti GPU. Power is supplied via four LiPo batteries housed in an impact-resistant battery cage, which also integrates Battery Management System (BMS) modules for safe and reliable power management \cite{artemis_open_source}.

\subsection{Foot Design}
As a humanoid soccer athlete, effective interaction with the ball plays a critical role in overall game performance. The primary objective of the foot attachment is to enhance the robot’s kicking accuracy and success rate without compromising locomotion stability. 


    
    
    

\begin{figure*}
    \centering
    \includegraphics[width=\linewidth]{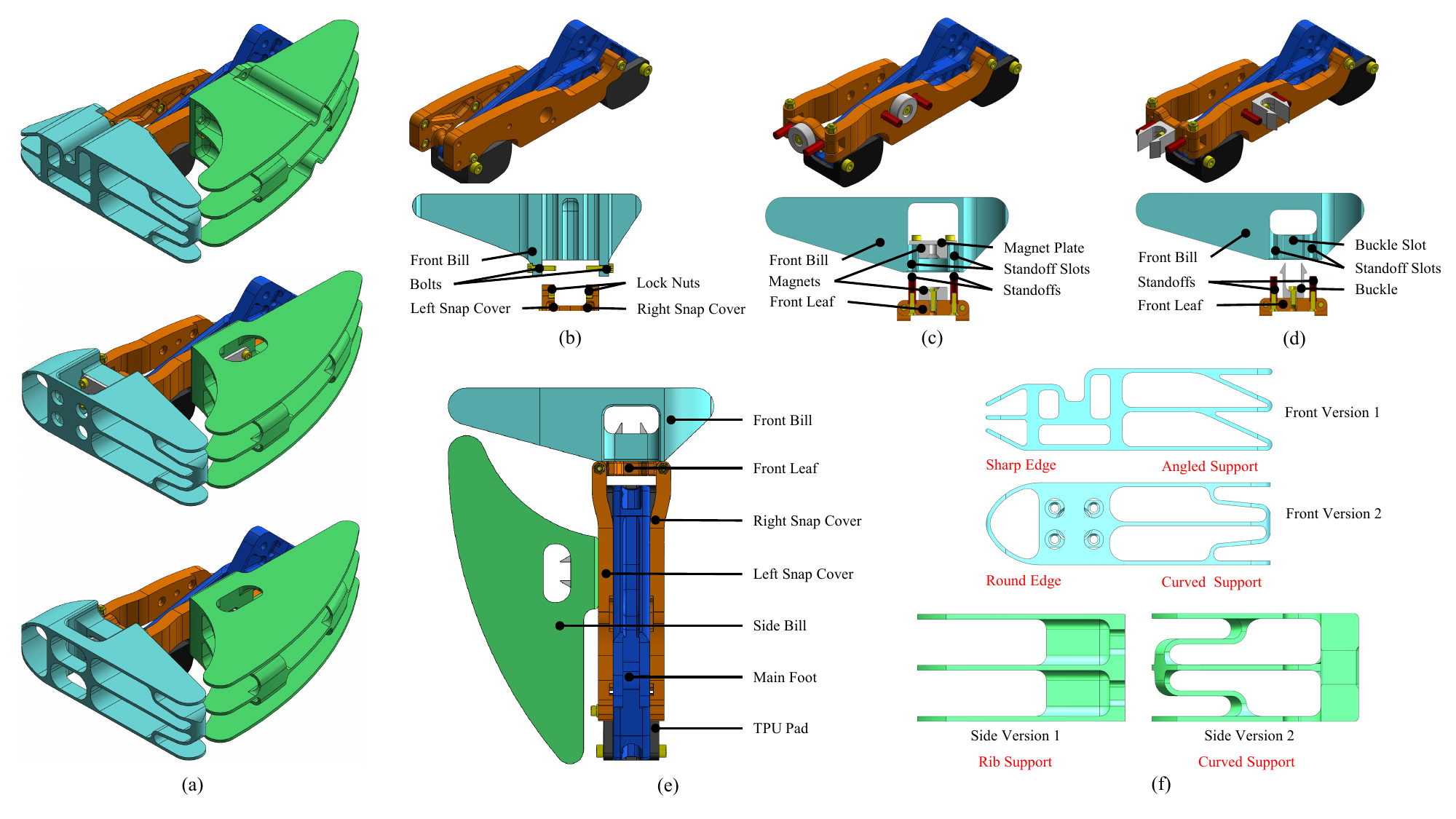}
    \caption{Foot Design a). Overview of three versions of design. b). Isometric view and sectional view of bolt version. c). Isometric view and sectional view of magnet version. d). Isometric view and sectional view of buckle version. e). Top view of selected (buckle) version assembly. f). Improvement of support structure for Front/Side Bill}
    \label{fig:Foot_Design}
\end{figure*}

As shown in \cref{fig:Foot_Design} (a), we designed and evaluated three versions of the foot attachment based on the requirements. \cref{fig:Foot_Design} (b)-(d) show the assembly without Front Bill and Side Bill. The main difference among each design lies in the interface between the Snap Cover and the Front/Side Bill. A detailed comparison is presented in \cref{table:foot_comparison} evaluating structural strength, connection robustness, replacement time, and cost. Based on the comparison, the third design was eventually selected as it achieved the best overall balance across evaluation criteria.


\begin{table}[h!]
    \centering
    \caption{Comparison of Three Connection Methods}
    \label{table:foot_comparison}
    \begin{threeparttable}
    \resizebox{\columnwidth}{!}{
    \begin{tabular}{ccccc}
        \toprule
        Design Version & Structural Strength & Connection Reliability & Assembly Time & Cost \\
        \midrule
        Bolt   & +++ & +++ & +   & ++   \\
        Magnet & +   & +   & +++ & +  \\
        Buckle & ++  & ++  & ++  & +++    \\
        \bottomrule
    \end{tabular}
    }
    \vspace{1mm}
    \begin{tablenotes}
        \footnotesize
        \item \parbox{\columnwidth}{“+” indicates relative performance level; more “+” symbols represent better performance in that category. For assembly time and cost, shorter assembly time and lower cost are considered better performance.}
    \end{tablenotes}
    \end{threeparttable}
\end{table}

The final foot attachment, shown in \cref{fig:Foot_Design} (e), consists of three main components: Snap Cover, which mounts to the foot’s main body and works as the connection with other components; Front Bill, which directly contacts the ball during forward kicks; and Side Bill, which aids in side kicking and prevents ball entrapment. This modular and anisotropic design provides both robustness and functional precision in high-dynamic, frequent-contact competition environments.





\subsubsection{Front and Side Bill Design}
As the primary component for ball contact, the Front Bill is designed to maximize the contact area at the leading edge, improving the consistency and accuracy of kicks. An asymmetric geometry is adopted to the part to minimize self-collision during turning and moving sideways. Additionally, the contact face is shaped as a flat plane to ensure the applied force is transmitted along the principal axis of the foot, enhancing energy transfer and yielding more predictable ball trajectories.

Complementing this design, the Side Bill is positioned to fill the gap behind the extended section of the Front Bill. The primary function is to prevent the ball becoming trapped within the gap and rolling backward during foot retraction. By closing this region, undesired contact with the ball is eliminated, enabling more reliable and predictable control of the kicking behavior. Additionally, the Side Bill provides a functional surface for side kicks when the robot is positioned near the goal, offering increased tactical flexibility in confined or angled scenarios.

The overall structure exhibit high stiffness in the horizontal direction, ensuring strong lateral force transmission during kicking, while maintaining compliance in the vertical direction. Furthermore, \cref{fig:Foot_Design} (f) shows the modifications we made to the structure. Based on the failure situations during the competition, the curved continuous supports are implemented for offering more uniform load distribution and reduced localized stress concentrations. Additionally, the inner edge was redesigned with a rounded profile, which increases structural stiffness and mitigates damage caused by edge impacts.



\subsubsection{Optimization for Quick Release Mechanism}
Based on the design requirements of the foot attachment, 3D printing was selected as the manufacturing method due to its rapid fabrication and geometric flexibility. However, the combined effects of material properties and the harsh operating environment inevitably lead to structural damage. To minimize maintenance time following failure, a buckle-based quick-release mechanism is incorporated into the design.


The front connection of the Snap Cover was redesigned as a dual-hinge mechanism, replacing the previous bolt interface. This configuration enables more accurate and less alignment during installation and requires tightening only a single bolt at the heel. In addition, a buckle–standoff scheme is employed to secure the Snap Cover to the Bills. In this design, the buckles lock the Front and Side Bills to their mounting surfaces, while laterally oriented standoffs provide structural support against vertical stepping loads, preventing cover detachment during collisions. This updated connection mechanism preserves replacement time of approximately 10 second, substantially faster than the fully bolted version, while withstanding nearly three times the detachment force of the magnet version.

\section{Software}
\subsection{Vision}
\label{sec:vision}

The vision system for the Robocup 2024 competition integrates the stereo vision from a ZED camera, customized deep learning models YOLOv8m based on YOLOv8 \cite{yolo_v8}, Kalman filter \cite{chen2018tracking} for tracking, and GPU optimization for fast cycle-times. Additionally, compared to traditional method on vision \cite{fiedler2019_humanoid_vision, he2022}, we employs both learning-based detection and classical computer vision to provide a fast perception system robust to environmental conditions and false-detections.



The system is built around a ZED stereo camera providing synchronized RGB and depth data, operating in \texttt{sl.DEPTH\_MODE.ULTRA} for high-precision depth estimation at 60 FPS with a wide field of view. Intrinsic parameters are dynamically adjusted based on the robot’s head orientation, with real-time retrieval of object positions from RGB and point cloud data, it enables accurate 3D projection and world coordinate transformation.

\subsubsection{Object Detection and Validation}

The core of object detection system is a custom-built version of the YOLOv8m model, which is reduced to approximately 80\% of the original size to increase the speed. The smaller network showed very minor decrease to its accuracy given our relatively small class size. The network is further optimized with NVIDIA TensorRT for GPU acceleration, achieving real-time high frame rate inference with minimal latency and a 70\% reduction in inference time compared to the standard YOLOv8m model.

The detection model is trained to recognize:
\begin{itemize}
    \item \textbf{Soccer Balls}: A primary object of interest, crucial for decision-making during gameplay.
    \item \textbf{Robots (Red and Blue Teams)}: The model distinguishes between robots from the red and blue teams based on their HSV color profiles.
    \item \textbf{Landmarks}: Features such as goalposts and intersections (T-, X-, or L-shaped) are detected to aid in navigation and localization.
\end{itemize}

Bounding boxes are computed for each detected object, along with confidence scores. Using depth information from the ZED camera's point cloud, these 2D detections are converted into 3D world coordinates. A cost function using the model confidence of a detection and its distance from the robot is used to provide the best 6 landmarks for localization. This conversion and calculation is essential for calculating the positions of the ball, robots, and best field landmarks.


As ball detection plays a critical role in the vision system and gameplay, the system employs several techniques to validate the ball’s detection:

\begin{itemize}
    \item \textbf{Green Coverage Validation}: Since the soccer ball is typically surrounded by green turf, the system verifies that a significant portion of the region around the detected ball contains green pixels. This validation step helps reduce false positives.
    \item \textbf{Dimension Constraints}: The system checks whether the dimensions of the detected object are within expected ranges for a soccer ball. Objects that do not meet these criteria are filtered out.
    \item \textbf{Height Constraints}: An additional filter that transforms the ball into world frame and checks its z-axis height is employed to add additional robustness to false detection. Given the assumption that the soccer ball will be on the floor, a threshold of 0.3 meters is used.
\end{itemize}

These validation steps ensure that the detected ball is correctly identified, even in cases of occlusion or partial visibility.

\subsubsection{3D Pose Estimation and Geometric Filtering}
For each detected object, the centroid of its bounding box is computed, and its 3D position is estimated by averaging or taking the median of the depth values within the bounding box. This process yields real-time, precise 3D coordinates for all detected objects.

In the specific case of ball detection, if depth data is unreliable or unavailable, the system employs a geometric depth estimation technique. This method uses the pixel diameter of the detected ball and the camera's intrinsic parameters to estimate the ball’s real-world distance from the robot. This approach serves as a backup to ensure continuous tracking of the ball, even in the absence of high-quality depth information.

\subsubsection{Kalman Filter for Object Tracking}
The vision system employs a Kalman filter for tracking objects, particularly the soccer ball. The Kalman filter is designed to handle noisy detections and predict future positions, allowing the robot to maintain accurate awareness of object locations even when detections are intermittent or unreliable. The filter tracks the position and velocity of objects in a six-dimensional state space, updating its estimates with every new detection.

The Kalman filter's predictive capabilities are particularly useful when objects are occluded or temporarily missed by the detector. By smoothing the detections and predicting the trajectory of objects, the Kalman filter ensures that the robot can respond proactively to the movement of the ball and other entities on the field.

\subsubsection{Real-Time Inference and Data Recording}


The vision system operates at 60 Hz, corresponding to the maximum camera polling rate in stand-alone mode, providing highly frequent updates of object positions and proximity estimates. To ensure optimal performance during gameplay, real-time inference results are logged and recorded, including raw RGB frames from the ZED camera and YOLOv8m outputs. The recorded data comprises bounding boxes, 3D coordinates, confidence scores, and depth information used for pose estimation. These logs are used for post-game diagnosis, model refinement, and performance evaluation, and the proximity data further enable detailed analysis of robot motion and obstacle avoidance behavior.
\subsection{Proximity}
\label{sec:prox}

Robust navigation and obstacle avoidance are critical in the highly dynamic competition environment, where physical entanglements with opponent robots frequently result in penalties or even falls. To address this challenge, we developed the Proximity module, a real-time depth perception system dedicated to close range obstacle detection and avoidance. The pipeline leverages on the stereo camera's depth data, executing a sequential pipeline to process a raw depth map into a set of actionable obstacle detections. The module primarily acts as a fallback for when vision-based detection fails during close contacts.

\subsubsection{Frame Preparation and Near-Field Collision Handling}
The pipeline begins by receiving a point cloud from the primary vision node where the depth map is extracted. To manage computational load, the frame is first downsampled.


The first analytical step addresses the intrinsic limitation of the camera, which cannot reliably measure depths closer than approximately 0.4 m and returns either $NaN$ or $Inf$ values. A dedicated check is implemented to analyze the density of these invalid measurements within the depth map. When the ratio exceeds a predefined threshold, we regard it as the presence of a very close object. The distance to this object is then estimated:

\begin{equation}
    d = d_{max} - \left(\frac{d_{max} - d_{min}}{c_{max} - c_{min}} \right) \times \left(c - c_{min}\right)
\end{equation}

where $d_{min}$ and $d_{max}$ are the minimum and maximum plausible distances, and $c_{min}$ and $c_{max}$ are the corresponding screen coverage percentages. A separate Connected Component Analysis (CCA) is running on this binary mask of special values to extract the obstacle's bounding box, which is appended to our list of detections with its interpolated depth.

\subsubsection{Geometric Field-of-View (FoV) Filtering}
\begin{figure} [htbp!]
    \centering
    \includegraphics[width=0.98\linewidth]{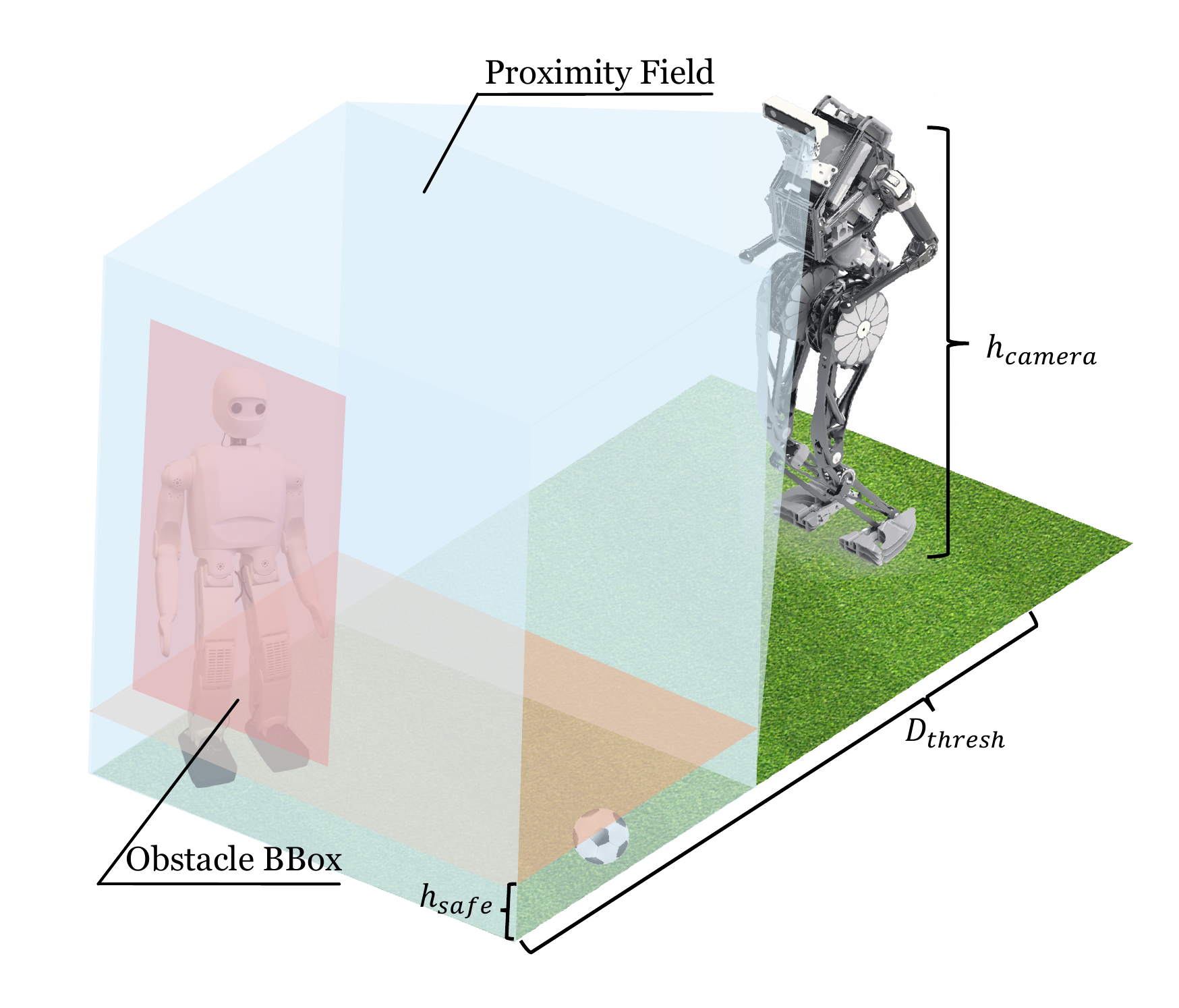}
    \caption{Visualization of the Geometric Field-of-View approach, filtering out expected depth map readings
and appending alien depth map values within the proximity field.}
    \label{fig:FOV-filter}
\end{figure}

After processing the near-field region, the raw depth map still contains information irrelevant to obstacle avoidance, including the ground plane and distant background structures. To retain only obstacles that physically intrude into the robot’s workspace, we apply a geometric filtering step (\cref{fig:FOV-filter}) based on the camera's field-of-view and the robot's head orientation.

Given the camera’s vertical field-of-view $\text{FOV}_v$ and the robot’s current pitch angle $\text{pitch\_rad}$, we compute the vertical ray angle associated with each image row $i$:

\begin{equation}
    \theta_i = \text{linspace}\!\left(-\frac{\text{FOV}_v^{\text{rad}}}{2}, \frac{\text{FOV}_v^{\text{rad}}}{2}, N\right)_i + \text{pitch\_rad}.
\end{equation}

Assuming a ground plane at height $h_{\text{safe}}$, and the camera at height $h_{\text{camera}}$, the depth where the ray would intersect the floor is:
\begin{equation}
    d_{\text{floor}}(i) = \frac{h_{\text{camera}} - h_{\text{safe}}}{\sin(\theta_i)}.
\end{equation}

Depth measurements greater than the value are discarded. We additionally enforce a global maximum interaction distance $D_{\text{thresh}}$ to ignore depth values associated with objects too far away to affect the robot’s immediate motion. The retained depth values therefore satisfy:

\begin{equation}
    0 < d(i,j) < \min\!\left(d_{\text{floor}}(i),\, D_{\text{thresh}}\right).
\end{equation}


This filter removes depth values associated with free space and distant background, retaining only measurements closer than the floor, indicating an object that protrudes upward into the robot’s traversable space. These retained values are treated as candidate obstacles and passed to subsequent processing stages.

\subsubsection{Obstacle Segmentation and Feature Extraction}
With a cleaned depth map, we employ CCA to segment the remaining regions into discrete objects \cite{HE201725}. Using an 8-connected neighborhood approach, the algorithm groups adjacent non-zero pixels into distinct components. We generate a bounding box for each valid component and then extract key features, including the mean depth of the pixels within the box and the box's coordinates in both the pixel frame and the world frame


\subsubsection{Bounding box refinement and Output}



The final stage refines the detected bounding boxes by merging overlapping or closely spaced boxes that likely correspond to the same physical obstacle. Sensor fusion is then performed by analyzing RGB pixel values within depth-consistent regions of each box to reclassify the robot’s team color, providing supplementary information to the primary vision system. The valid bounding boxes are subsequently sorted by mean depth, and the two closest obstacles are published to the navigation and behavior modules. If fewer than two objects are detected, dummy entries are inserted to maintain a consistent data structure.





\subsection{Localization}
\label{sec:localization}



Localization in RoboCup is essential: each robot must know its position $(x,y)$ and orientation $\theta$ to decide whether to shoot, pass, navigate to a set-play location, or face the correct goal. Because RoboCup limits sensing to human-analog sensors (camera and IMU), localization must rely on noisy, intermittent visual and inertial data in a dynamic environment with moving robots, occlusions, changing lighting, camera flashes, and motion blur. The system must also run with low latency, as slow or unstable estimates directly degrade path tracking and can cause collisions; in practice, path-planning performance is tightly coupled to localization delay \cite{2025_hou_clap_localization}.

CLAP \cite{fernandez2025clap}, uses a lightweight geometric strategy that produces state hypotheses which naturally cluster when camera observations align with the known field map. The algorithm selects the pose whose predicted landmark positions best match the detections. Its simplicity and parallelism enable fast and accurate localization. Clustering in both the robot and field frames filters outliers and improves robustness to misdetections, while fusion with inertial measurements maintains a continuous pose between camera updates.

\subsubsection{Landmark Matching}
\label{sec:matching}
Because most soccer-related behaviors depend only on the robot’s planar pose $(x_{\text{robot}}, y_{\text{robot}}, \theta_{\text{robot}})$ with respect to a field coordinate frame, the localization task can be treated as a 2D problem by projecting all camera-detected landmarks onto a plane. \cref{fig:cover} (b) presents the set of recognizable field features, including corners (\textit{L}), goalposts (\textit{G}), T-junctions (\textit{T}), and crosses (\textit{X}). Each landmark is encoded by its class label and its relative position to the robot, without estimating any orientation. It is important to note that \cref{fig:cover} (b) illustrates an idealized situation in which features are observed perfectly, with no positional deviation (what we refer to as noise) and without any missed or spurious detections, collectively described as false detections.

Given the a pair of observed field features relative to the robot frame ($p_1^{\text{body}}$, $p_2^{\text{body}}$) and a set of matching features in world frame ($p_1^{\text{world}}$, $p_2^{\text{world}}$) from the a priori known map where the landmark orderings are the same (e.g.  \( p_1^{\text{world}} \) and \( p_1^{\text{body}} \) are both \textit{T}s), the robot's orientation \( \theta_{\text{robot}} \) can be calculated as:
\begin{equation*}
   \theta_{\text{robot}} = \text{atan2} (\Delta p_{y}^{\text{world}}, \, \Delta p_{x}^{\text{world}}) - \text{atan2} (\Delta p_{y}^{\text{body}}, \, \Delta p_{x}^{\text{body}}) 
\end{equation*}
\begin{equation}
    p_{robot} =\begin{bmatrix}
        x_{\text{robot}} \\ y_{\text{robot}}
    \end{bmatrix} = p_{1}^{\text{world}} - R(\theta_{\text{robot}})*p_1^{body}
\label{eq:mirror_match}
\end{equation}
Where $\Delta p = p_2 - p_1$, $p_{\text{robot}}$ is the robot's global position, and $R$ is a 2D rotation matrix.

When more than two landmarks are observed, every pair of landmarks can be used to generate a pose hypothesis, producing a set of candidate robot states. However, a single landmark pair often corresponds to multiple valid matches on the field. This ambiguity arises from several factors: the geometric symmetry of a standard soccer field, identical landmark labels that cannot be distinguished by appearance alone, and measurement noise that makes the distances between certain landmark pairs difficult to differentiate. As a result, each observed pair may map to multiple possible field locations. To avoid prematurely discarding feasible solutions, all matches whose distances fall within a predefined tolerance are accepted, and the corresponding pose hypotheses are added to the candidate set for later clustering.

\subsubsection{Clustering}
\label{sec:clustering}


The landmark matching process in \cref{sec:matching} generates many pose hypotheses, most of which are incorrect. Spurious candidates are widely scattered due to symmetry, ambiguous associations, and noise, while the true pose forms a dense cluster. This motivates clustering for state estimation. CLAP therefore uses two complementary strategies: a fast local clustering method exploiting temporal consistency, and a global clustering method that aggregates candidates over multiple frames for verification and recovery.

\begin{figure}[htbp!]
    \centering
    \includegraphics[width=\linewidth]{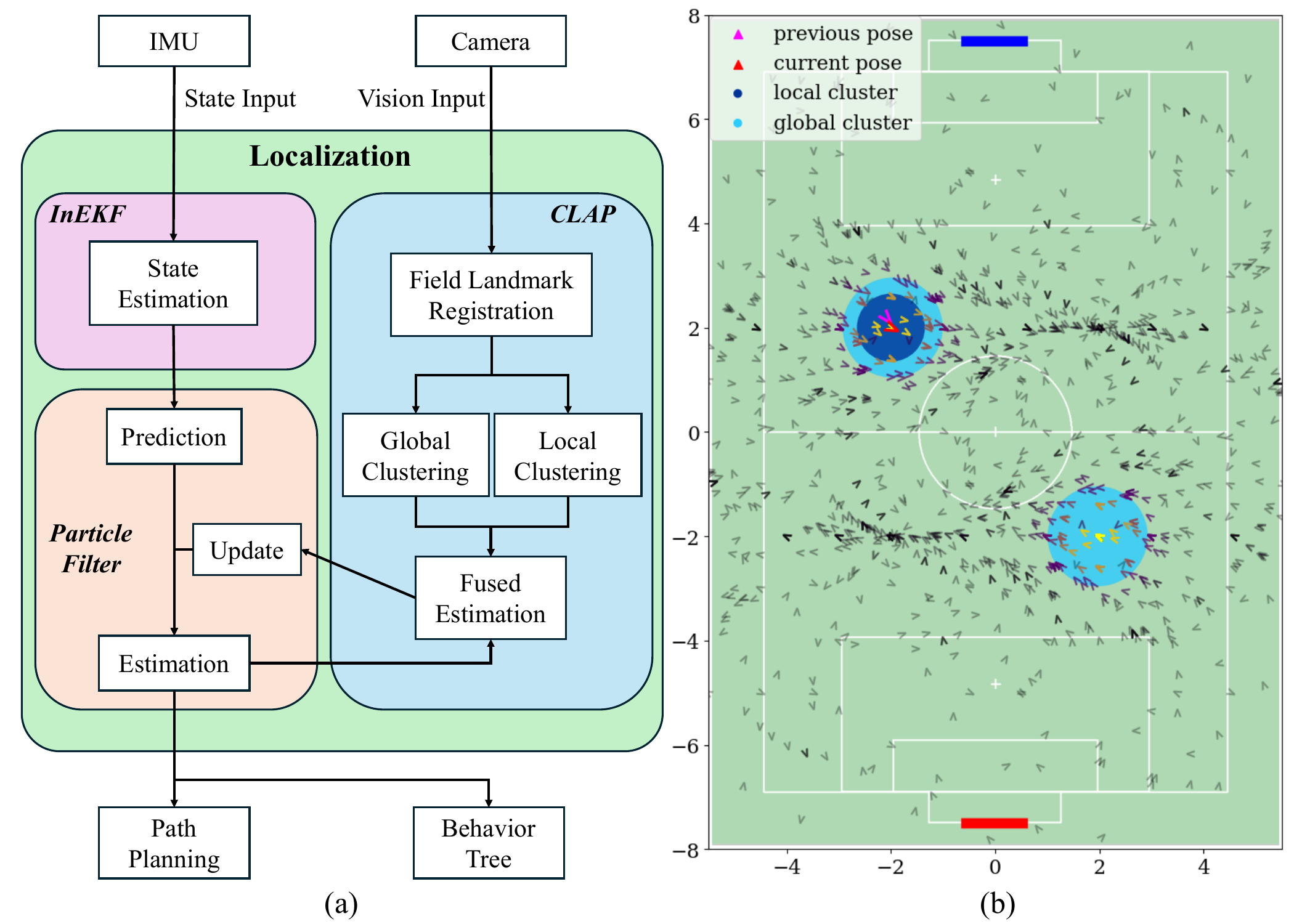}
    \caption{Localization method. a). Localization framework overview. b). Probability distribution of localization cluster.}
    \label{fig:localization_method}
\end{figure}

To reduce computation and improve update speed, a simple local clustering strategy is used that evaluates only the cluster located within a fixed radius of the previous pose estimate, shown by the dark blue circle of \cref{fig:localization_method} (b). Despite its simplicity, this approach performs reliably; even under significant noise, the estimate often returns to the correct region, forming a broad ``bowl of attraction.'' This basin can be observed in \cref{fig:localization_method} (b) from the lighter-colored arrows surrounding the deepest point, highlighted by the yellow arrow. The next pose is obtained by averaging the estimates within the chosen cluster, so if the cluster is offset from the center of the bowl, surrounding estimates naturally pull it toward the correct location.

For increased robustness, a slower global clustering procedure also aggregates state estimates over multiple frames, supported by IMU predictions whose drift remains small over short intervals. This global method periodically checks the na\"ive estimate, and if the two differ beyond a set threshold, the pose is reset to the global result—represented as the \textit{Fused Estimates} step in \cref{fig:localization_method} (a). Due to the symmetry of the soccer field, this technique can typically reduce the ambiguity to two mirrored positions; with limited observations, it may produce several possible states. To resolve this, the centroid nearest to the previous estimate is selected.  
Because many estimates may be outliers, raw $K$-means centroids can be biased. To improve robustness, the algorithm iteratively applies $K$-means while removing a fraction of outliers after each iteration. This refinement worked well in practice. More details of how clustering is incorporated into CLAP is provided in \cite{fernandez2025clap}.

\subsection{Navigation}
\label{subsec:navigation}

\begin{figure}[t!]
    \centering
    \includegraphics[width=\linewidth]{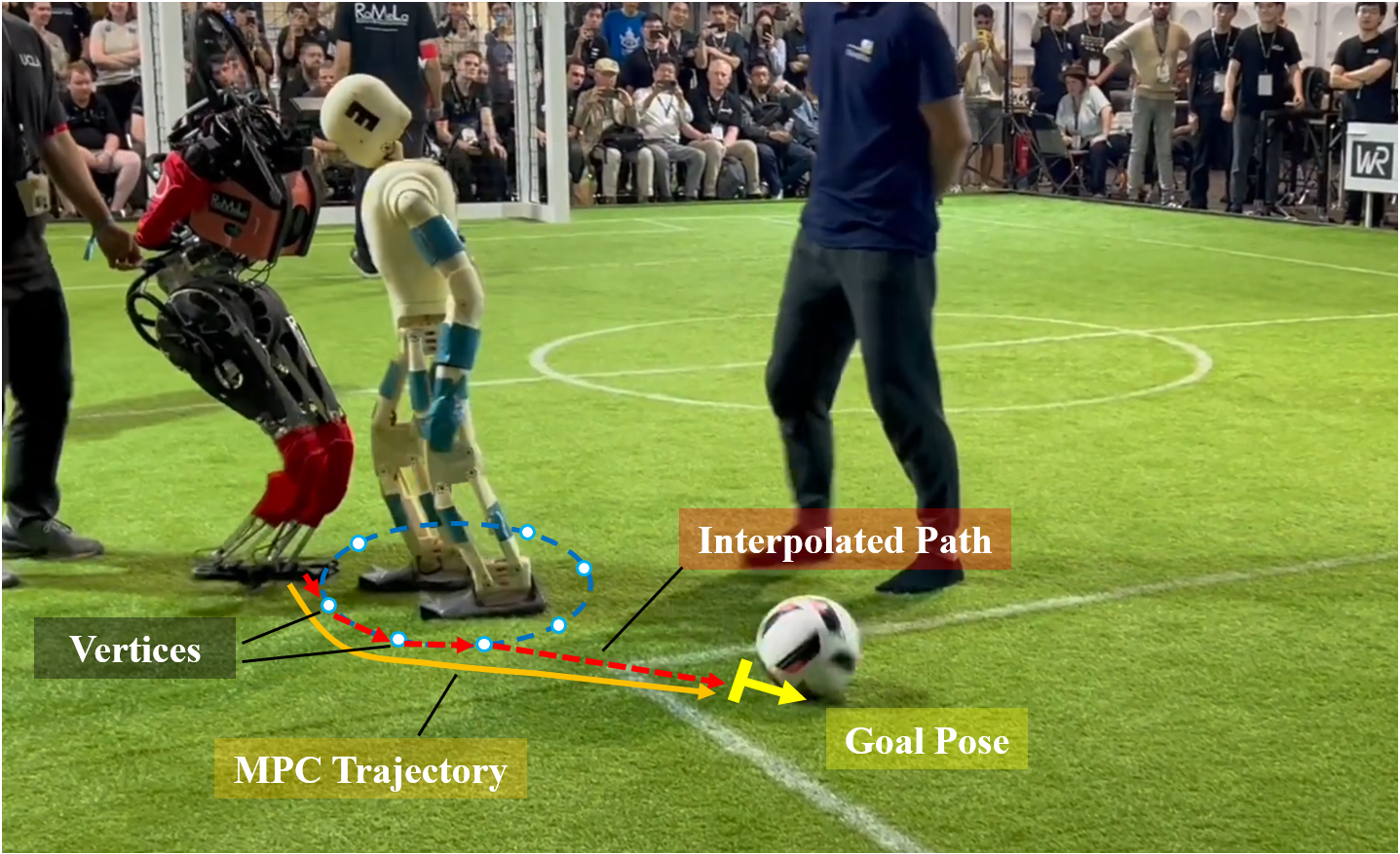}
    \caption{Figure shows a situation ARTEMIS in red is trying to avoid the opponent in blue to get to the ball during the championship match in RoboCup 2024.}
    \label{fig:midlevel_situation}
\end{figure}

We use a two-stage navigation stack that couples a fast, turn-aware global planner with a predictive tracker that enforces collision safety without mode switches as seen in \cref{fig:midlevel_situation}. DAVG planner computes a sparse route that prefers gentle turning when it is advantageous for a humanoid’s maneuverability, while cf-MPC follows that route while balancing dynamic feasibility and obstacle avoidance within a single optimization problem \cite{hou2025icra}.

\subsubsection{Planning via DAVG}
Obstacles (humans/robots) are modeled as regular polygons, and a visibility graph is built over the start, goal, and non-occluded obstacle vertices. To avoid searching the entire field at every replan, DAVG restricts computation to an \emph{active region} around the start–goal (S–G) line: obstacles intersecting the S–G line seed the region; the farthest relevant vertices along and perpendicular to S–G bound it; and the region expands iteratively to include any newly intersecting obstacles. This pruning retains only obstacles that can influence the route, reducing the number of nodes and visibility checks. When the start or goal happens to lie inside an obstacle footprint, we attach an “exit/entry” edge to the nearest feasible vertex in the heading/approach direction so that a valid path can be constructed immediately.

DAVG then augments the classical visibility graph with a turn cost. Instead of weighting edges purely by Euclidean length, we define edges over \emph{ordered} vertex pairs and penalize the incremental heading change required to transition from one segment to the next. This combined metric preserves the desirable structure of visibility-graph shortest paths while biasing solutions away from sharp cornering that is costly for a large humanoid. The resulting waypoints are time-stamped by $C^0$ (piecewise-linear) interpolation, yielding a simple, replan-friendly reference trajectory $X_r$ for the tracker \cite{hou2025icra}.

\subsubsection{Tracking via cf-MPC}

To track the DAVG reference while avoiding obstacles, we use Collision-Free MPC (cf-MPC). The robot is modeled as a 3-DOF planar system with state $X=[x,\,y,\,\theta]^\top$ and body-frame velocity input $u=[v_x,\,v_y,\,\omega]^\top$. Using a body-frame integrator, the discrete dynamics are
\begin{equation}
\begin{bmatrix} x_{k+1}\\ y_{k+1}\\ \theta_{k+1}\end{bmatrix}
=
\begin{bmatrix} x_k\\ y_k\\ \theta_k\end{bmatrix}
+
\begin{bmatrix}
\cos\theta_k & -\sin\theta_k & 0\\
\sin\theta_k & \cos\theta_k  & 0\\
0 & 0 & 1
\end{bmatrix}
\begin{bmatrix} v_{x,k}\\ v_{y,k}\\ \omega_k\end{bmatrix} dt.
\end{equation}

Over a horizon $N$, cf-MPC solves a nonlinear MPC problem that jointly penalizes tracking error, control effort, and constraint violation:
\begin{equation}
\min_{\{X_i,u_i,\delta_{i,j}\}}\;
\sum_{i=1}^{N} \|X_i - X_{r,i}\|_{Q}^{2}
+
\sum_{i=0}^{N-1} \|u_i\|_{R}^{2}
+
\rho \sum_{i,j} \delta_{i,j}^{2},
\end{equation}
subject to the dynamics, input bounds on $(v_x, v_y, \omega)$, and soft circular keep-out constraints for each obstacle $j$:
\begin{equation}
(x_i-x_j)^2 + (y_i-y_j)^2 + \delta_{i,j} \;\ge\; R_j^{2}, 
\qquad \delta_{i,j} \ge 0.
\end{equation}
The slack variables $\delta_{i,j}$ preserve feasibility when obstacles suddenly intrude or estimator noise momentarily places the robot inside a keep-out set, while the single objective naturally balances trajectory tracking and obstacle avoidance without any hand-tuned mode switching.

In addition to this nonlinear formulation, we also employ a linearized quadratic-program (QP) variant of cf-MPC that approximates the dynamics and constraints around the current reference to achieve higher control rates; full derivations and implementation details are provided in our prior work~\cite{hou2025icra}.

\subsection{Behavioral Planner}
\label{sec:BT}

\begin{figure*}[htbp!]
    \centering
    \includegraphics[width=\linewidth]{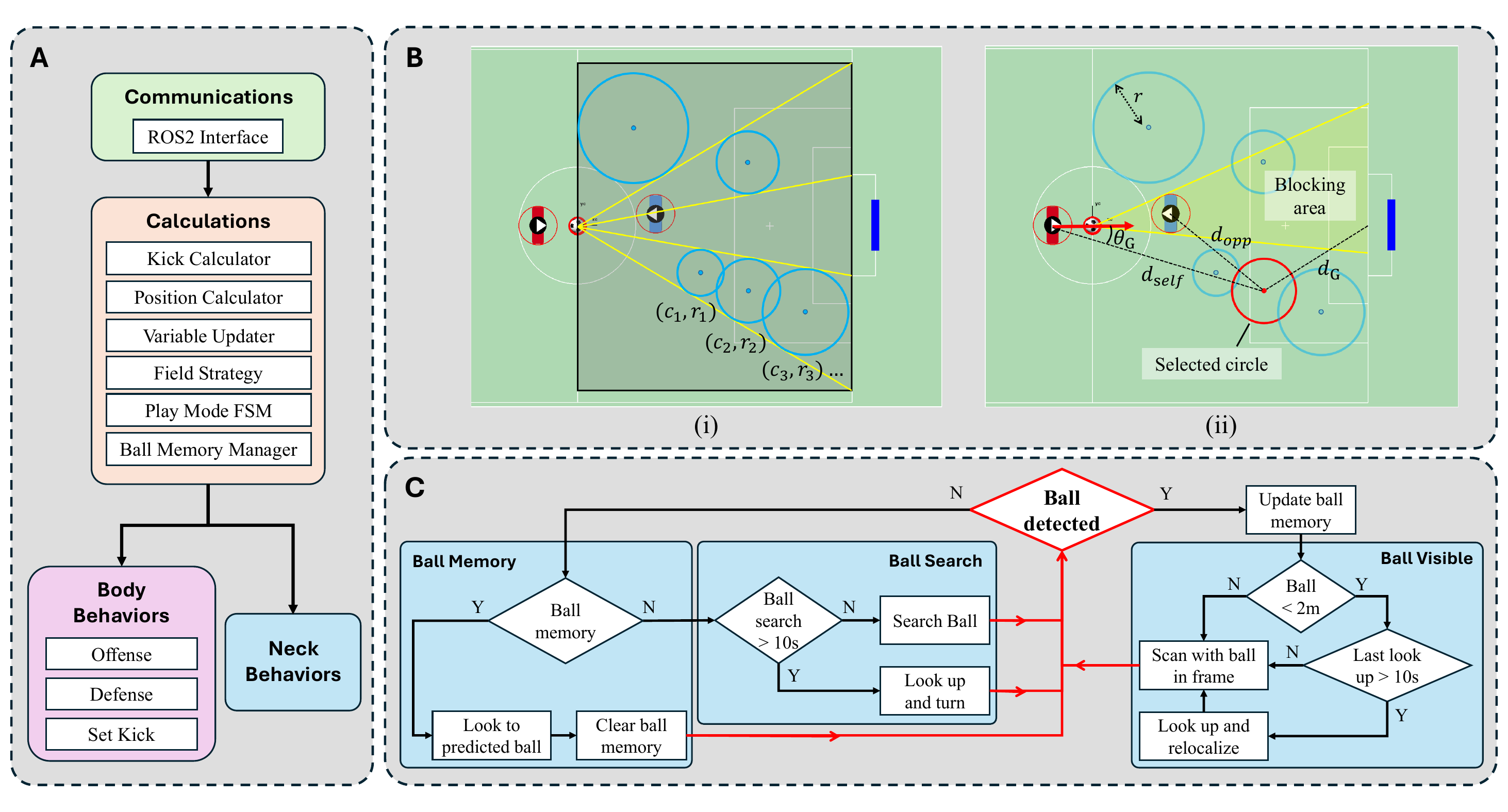}
    \caption{A: Overview diagram showing the components within the high-level behavior-tree. B: A demonstration of the pass algorithm: (i) visibility rays and field
boundaries generate candidate pass circles; (ii) each circle is evaluated using goal angle, teammate accessibility, and opponent-blocking metrics, and the best-scoring circle is selected as the pass target. C: Neck behavior logic flow: depending on visibility and memory, the neck switches between states to maintain robust localization and ball awareness.}
    \label{fig:High_Level}
\end{figure*}

When operating during the competition, decisions must be made almost instantaneously to deal with continually changing environment. We employed a hybrid behavior tree with FSM approach to tackle this problem. The system separates strategic decisions (e.g., whether to kick) from more directional directives (e.g., where and how hard to kick) and communications to other modules. As shown in \cref{fig:High_Level}(a), this is achieved through three segments, Body and Neck Behaviors (Decisions), Calculations, and Communications. The approach retains behavior-tree simplicity, adds FSM logic without latency, and uses parallelization to ensure proper sequencing of time-critical actions. The Communications segment uses ROS2 to receive messages from other components and transmit instructions to the corresponding nodes. The other two modules are described in detail below.


\subsubsection{Calculations}
The Calculations segment operates in parallel with the Decisions segment, handling complex determinations so that Decisions can remain lean and run efficiently without repeating expensive calculations. A central component is the positioning behavior, which determines the optimal pose for the intended action and generates kick commands. Lacking direct locomotion feedback, it predicts the robot’s future position to time kicks between strides. Kick power follows a distance-based model, with goal shots always taken at maximum power.


The ball’s target location is chosen from two modes: shooting or passing. Passing occurs when a direct shot is blocked or demands excessive kick accuracy. The pass-finding algorithm in the top-right of \cref{fig:High_Level} selects an open region by reducing the continuous space of possible targets to a finite set of candidates. These are created by drawing a series of bounding lines running through the current position of the ball, $\mathbf{b} = (b_x, b_y)$, and key points such as the corners of the goal and field as well as running tangentially to circles representing opposing players, $\{\mathbf{o}_1, \mathbf{o}_2, \ldots, \mathbf{o}_N\}$. The goal lines of the field and borders offset from the sidelines are added to this set. Triplets of these rays are checked for valid triangles and are used to generate candidate circles using a weighted centroid:

\begin{equation}
\mathbf{c} = \frac{a \cdot \mathbf{p}_1 + b \cdot \mathbf{p}_2 + c \cdot \mathbf{p}_3}{a + b + c}
\end{equation}

and Heron's Formula:
\begin{equation}
r^2 = \frac{(s-a)(s-b)(s-c)}{s}
\end{equation}

where $s = (a + b + c) / 2$ is the semi-perimeter.

From here we apply a weighting heuristic with consideration to the candidates distance from and angle to the goal, distance from ourselves, our teammates, and our opponents, and the size of the candidate which represents the required accuracy for the kick for each candidate circle $\mathbf{c} = (c_{xy}, r^2)$ which are stored within $C = {(c_j, r_j)}$. Each candidate center is scored with a scalar cost:

\begin{equation}
    \begin{aligned}
        J(c) =\;& 
        w_{\text{a}}\, J_{\text{area}}(c) 
        + w_{\theta_G}\, J_{\theta_G}(c)
        + w_{d_G}\, J_{d_G}(c)
        + w_{\text{s}}\, J_{\text{self}}(c) \\
        &\quad + w_{\text{t}}\, J_{\text{team}}(c)
        + w_{\text{o}}\, J_{\text{opp}}(c)
        + w_{\text{b}}\, J_{\text{block}}(c)
    \end{aligned}
\end{equation}

where
\begin{itemize}
    \item $w$ are the weight;
    \item $J_{area}(c)$ is the cost of circle dimension, which favors larger circles which provide higher tolerances for a successful kick;
    \item $J_{\theta_G}(c)$ is the cost of kick angle, which minimizes the amount of reorienting to the desired kick location;
    \item $J_{d_G}(c)$ is the cost of distance to the goal, which favors candidates that place the ball as close to the goal as possible;
    \item $J_{\text{self}}(c)$, $J_{\text{team}}(c)$, $J_{\text{opp}}(c)$ are the cost of the distance to the robot, teammate, or opponents, which favor circles close to the teammate and penalize ones close to opponents;
    \item $J_{\text{block}}(c)$ is the cost of clean trajectory, which penalizes candidates that have opposing players in or near their trajectories.
\end{itemize}

These are weighted and combined to create unique strategies for parametrically choosing one of the generated candidate positions to kick the ball to. The candidate is selected based on

\begin{equation}
    (c^\star, r^\star)
    =
    \arg\max_{(c,r)\in\mathcal{C}} J(c),
\end{equation}

where $c^\star$ is the pass target, and two tangents from the ball to the circle of radius $r^\star$ define the admissible pass cone and its angular tolerance for the kicking controller.

Another behavior we will address is the Action Mode FSM Behavior. This relates to translating the general output of the Decision Tree into discrete parameters such as movement codes which will be sent to the navigation node. Parameters regarding our play strategy such as how to split the playing field between our two players or which kick selection strategy to employ can also be adjusted autonomously based on the state of the game.

The third major behavior handles the system’s memory of the field and object permanence. From a high level perspective, we categorized four primary ball conditions from which we could prescribe different actions. The state of the ball is described as either Lost, In-View, Remembered, or Predicted. The ball remains in the In-View state when it is visible, and  transitions to Remembered once it leaves the camera’s field of view. The last known position is retained until it reappears, remains invisible for too long, or is expected to be visible based on the robot’s pose. Occlusions are handled by projecting shadows behind obstacles, preventing premature clearing of the ball memory. This heuristic reduces unnecessary search behavior and improves overall efficiency.


\subsubsection{Body and Neck Behaviors}
After completing these calculations, the Tree will move on to the Decisions section. This section is first divided into two parallel trees, the Body Behavior Tree and the Neck Behavior Tree. 

The Body Tree interprets incoming signals to select the robot’s Mode, then branches into Offense or Defense and, when applicable, into game-specific Set Piece states such as kickoffs or corner kicks. To maintain clarity and avoid excessive complexity, larger behaviors are modularized into reusable action blocks, including searching for the ball, kicking, and navigating to set points.

The Neck Tree operates in parallel with the Body Tree and manages all head-orientation behaviors according to the current game mode. Isolating neck control keeps the architecture simple, allowing frequent, short-horizon adjustments—such as those triggered by ball memory updates or de-localization risk—without complicating body behaviors. Depending on ball visibility, the tree shifts between tracking, prediction, search, and re-localization routines, as illustrated in the lower-right panel of \cref{fig:High_Level}.



\subsection{Locomotion} 
\label{sec:Locomotion Method}

\subsubsection{Locomotion Overview}
The locomotion of \textit{ARTEMIS} builds on a hierarchical, model-based control framework previously demonstrated in~\cite{artemis_open_source,ahn2023_thesis}. A top-level gait generator synchronizes leg contacts and provides phase variables to downstream planners. Robot state is estimated with a contact-aided Invariant Extended Kalman Filter (InEKF) and contact estimator~\cite{2019_InEKF,colin_thesis}. A footstep planner then selects the next stabilizing foothold from the current CoM state and desired velocity that minimizes predicted angular-momentum error about the stance foot~\cite{angular_momentum}.

The resulting target footholds are forwarded to the trajectory planner, which generates smooth swing motions using sixth\text{-}order polynomial trajectories. To accommodate kicking behaviors, the mid\text{-}swing waypoint of the trajectory is exposed to kick commander, which can adjust this intermediate point to shape the desired kicking motion. All swing trajectories are parameterized in a normalized phase space ($\varnothing \in [0,1]$).

At the control level, a weighted Whole-Body Controller (WBC) implicitly enforces task hierarchies across all limbs. Subject to the rigid-body dynamics and contact constraints. The WBC outputs joint torque commands, and then tracked by a low-level joint torque controller. This architecture allows ARTEMIS to stably walk up to $1.2$~m/s while maintaining balance over variable terrains.

\subsubsection{Perception-Locked Mid-Swing Kicking}
\begin{figure}[htbp]
  \centering
  \includegraphics[width=0.95\columnwidth]{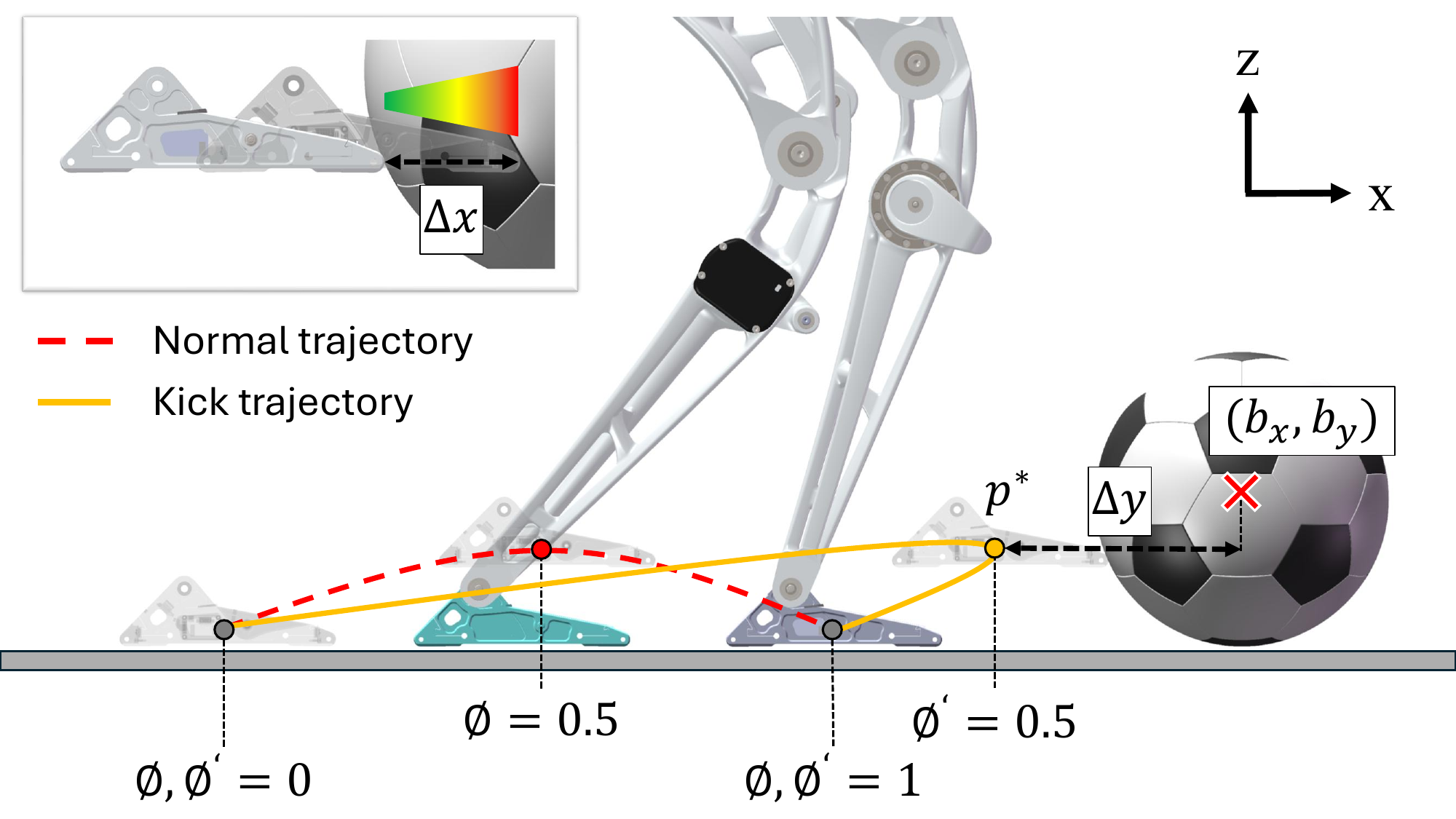}
  \caption{Kick geometry and timing diagram illustrating the left-leg swing phase for both the normal trajectory (dashed red) and the kick trajectory (solid yellow). The right leg (blue-green) remains stationary during the maneuver. The geometric offset is denoted by $\Delta y$, while $\Delta x$ adjusts the kick power.}
  \label{fig:kick_geometry}
\end{figure}

While locomotion provides dynamic stability, soccer performance requires rapid, accurate ball interaction without interrupting gait. To achieve this, a perception-locked mid-swing kicking (PLMK) module was integrated into the locomotion stack, enabling autonomous kicks executed within a normal walking step. \cref{fig:kick_geometry} illustrates the mid-swing retargeting of the swing foot toward the perceived ball position.

The kicking decision is governed by a \texttt{KickCommander} node,
which evaluates ball perception, gait phase timing, and internal re-arming conditions. The complete decision process is summarized in
Algorithm~\ref{alg:plmk}. At every control cycle, the commander first checks a set of locking conditions: if the robot is currently balancing, if either foot is already engaged in a kick, if no fresh ball measurement is available, if the forward velocity drops below $-0.1$~m/s, if the ball lies outside the in-kick zone $\mathcal{Z}$, or if fewer than three alternating touchdowns have occurred since the previous attempt, then the module exits without issuing a command. 

Once all conditions are satisfied, the system determines which foot should act as the striker based on the lateral ball position $b_y$: a ball on the left ($b_y>0$) arms the left foot, while a ball on the right ($b_y\le 0$) arms the right foot. When the selected foot enters swing (liftoff), the corresponding kick is initialized by storing the latest ball pose $b$, computing the desired strike point $p^\star$ in the foot frame, and marking the kick as active. After
execution, the kick is terminated on touchdown of the striking foot, clearing the in-progress flags and resetting the re-arm counter.

\begin{algorithm}[t]
\caption{Perception-Locked Mid-Swing Kicking (PLMK)}
\label{alg:plmk}
\begin{algorithmic}[1]
\Require latest ball pose $b$, forward speed $v_x$, touchdown log $\mathcal{T}$
\Ensure request flags $r_L,r_R$, in-progress flags $p_L,p_R$
\If{balancing $\lor$ $(p_L\lor p_R)$ $\lor$ no fresh $b$ $\lor$ $v_x < -0.1$ $\lor$ $b \notin \mathcal{Z}$ $\lor$ touchdowns $<3$}
  \State \Return
\EndIf
\State $r_L \gets (b_y > 0)$;\quad $r_R \gets (b_y \le 0)$
\Statex \textbf{On liftoff of foot $f \in \{L,R\}$ and $r_f$:}
\State $b^\star \gets b$;\quad compute strike $\mathbf{p}^\ast$ in foot-$f$ frame
\State \textit{aimed}$\gets$true;\quad $p_f\gets$true
\Statex \textbf{On touchdown of foot $f$:}
\State $r_f\gets$false;\quad $p_f\gets$false;\quad reset re-arm counter
\end{algorithmic}
\end{algorithm}

At liftoff, the perceived ball position $p_\mathrm{ball}$ is projected into the kicking-foot frame and adjusted by a geometric offset $\Delta y$ and kick power scaling $\Delta x$ to define the strike point:
\begin{equation}
\mathbf{p}^\ast = \mathrm{clip}_{\mathcal{B}}\!\left( R_z(\psi_f)^{\!T}(p_\mathrm{ball}-p_\mathrm{stance})
+ [\Delta x(\mathrm{power},r), \, \Delta y]^\top \right),
\end{equation}
The target point $\mathbf{p}^\ast$ is passed to the swing-leg trajectory solver as a mid-swing constraint:
\begin{equation}
p_\mathrm{foot}(t_m) = \mathbf{p}^\ast,
\end{equation}
forcing the sixth-order planar swing polynomial to intersect the ball at mid-swing. 
This ensures impact occurs when the leg is at maximum velocity while preserving smooth gait transitions.

\subsubsection{System Integration and Strategic Benefits}
The PLMK module interfaces directly with the swing-leg trajectory generator inside the whole-body control pipeline, operating fully autonomously from perception through execution. Ball detections and depth estimates from the camera are fused with localization outputs to provide global robot and ball poses for accurate relative targeting. These estimates feed into the strategy planner, which selects between passing, dribbling, or shooting by adjusting the kick power. Once a kick is issued, locomotion incorporates the motion seamlessly into the ongoing gait cycle, enabling ARTEMIS to strike the ball while walking. By mapping kick power to tactical behaviors such as short passes, controlled dribbles, or long-range shots, the robot can adapt its play style dynamically in real time.

\section{Results}
Our proposed system was evaluated through in-field subsystem verifications, simulations, and full stack soccer matches. The subsystem verifications demonstrated that the perception, locomotion and kicking, localization, works well. We further analyzed the foot attachment material and high-level strategy in simulation, and further we evaluated our full stack soccer performance through pseudo and actual soccer matches, demonstrated the robustness, component integration works well with the strong soccer performance. 

\subsection{In-Field Verification of Core Subsystems}
This section evaluates the core subsystems of our soccer architecture through a sequence of controlled in-field tests. We begin by isolating perception modules—including vision, proximity sensing, and localization—to verify their reliability under realistic on-field conditions. Building on this foundation, we assess sub-layer integrations by validating reference trajectory tracking, followed by combined perception and mid-level path planning with closed-loop tracking. Finally, we evaluate kicking performance in conjunction with perception and locomotion. Together, these staged subsystem evaluations provide a structured verification that each layer of the pipeline functions robustly on the physical field, enabling the dynamic soccer performance demonstrated in later sections.

\begin{figure*}[htbp!]
    \centering
    \includegraphics[width=\linewidth]{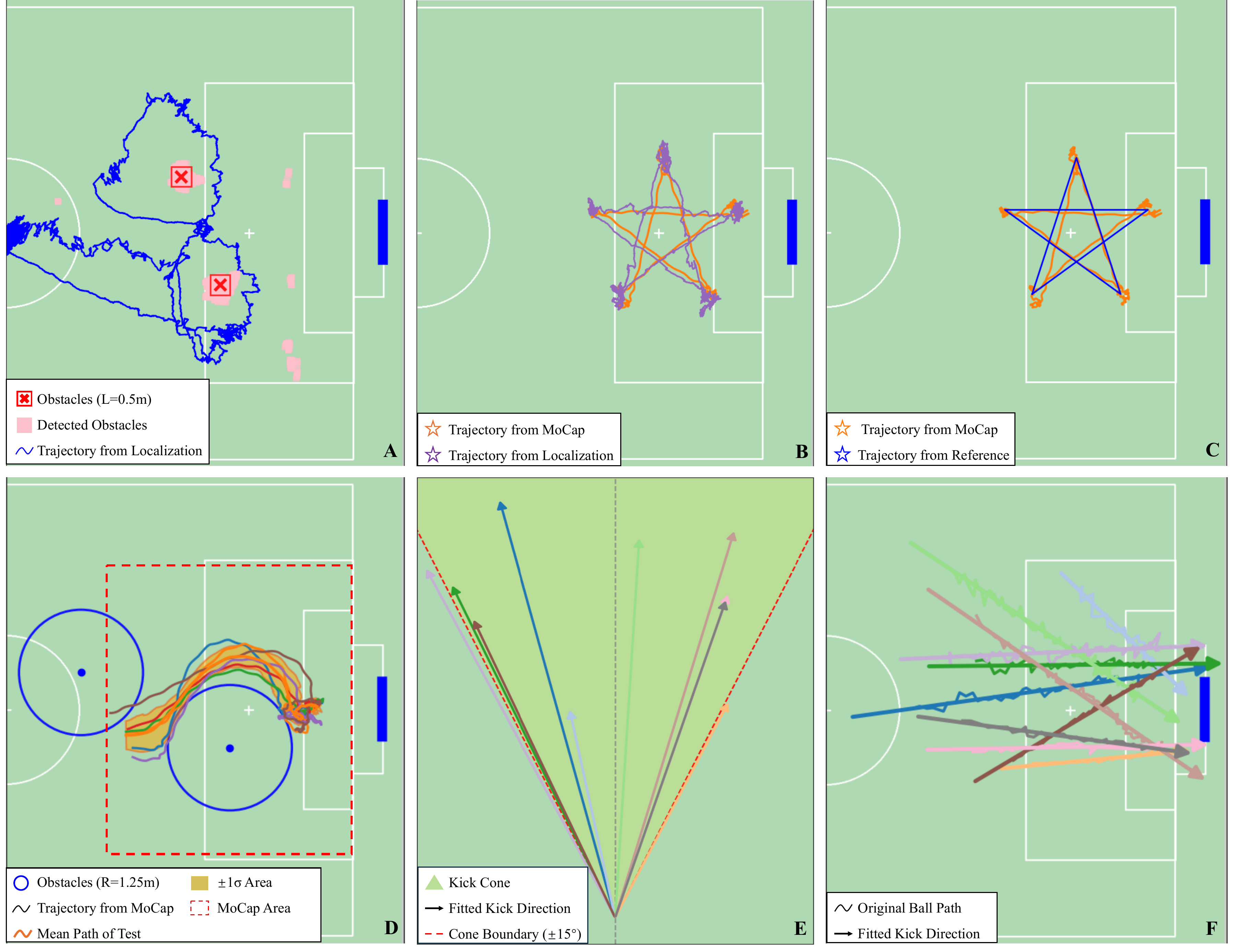}
    \caption{In-Field verification of Core Subsystems Result. A: Obstacle detection result from proximity; B: Trajectory comparison between localization and motion capture; C: Trajectory comparison between mid-level tracking and reference trajectory D: Navigation test with obstacle avoidance.E: Kick direction and preset kick cone comparison; F: Ball path from localization and fitted kick direction.}
    \label{fig:subsystem result}
\end{figure*}

\subsubsection{Perception}
To assess the performance of the detection model, we adopt standard object detection metrics, namely Precision (P), Recall (R), and mean Average Precision (mAP). These metrics are defined as follows:

\begin{itemize}
    \item \textbf{Precision (P):} Measures the proportion of correctly predicted positive detections out of all predicted positives. 
    \[
    P = \frac{TP}{TP + FP}
    \]
    where $TP$ is the number of true positives and $FP$ the number of false positives.

    \item \textbf{Recall (R):} Measures the proportion of correctly predicted positive detections out of all actual positives. 
    \[
    R = \frac{TP}{TP + FN}
    \]
    where $FN$ is the number of false negatives.

    \item \textbf{Average Precision (AP):} For each class, precision is plotted against recall at varying confidence thresholds. The area under this curve yields the Average Precision (AP). 

    \item \textbf{Mean Average Precision (mAP):} The mean of AP scores across all classes. We report:
    \begin{itemize}
        \item \textbf{mAP@50:} AP computed at an Intersection-over-Union (IoU) threshold of 0.5.
        \item \textbf{mAP@50:95:} AP averaged over multiple IoU thresholds from 0.5 to 0.95 with a step size of 0.05, following the COCO evaluation protocol.
    \end{itemize}
\end{itemize}


\begin{table}[htbp]
\centering
\caption{Benchmark results of the custom YOLOv8m model on a test set comprised of $350$ images taken from challenging points of views from the 2024 RoboCup official field. Metrics include Precision (P), Recall (R), mAP@50, and mAP@50:95.}
\label{tab:vision_results}
\begin{tabular}{lcccc}
\hline
\textbf{Class} & \textbf{P} & \textbf{R} & \textbf{mAP@50} & \textbf{mAP@50:95} \\
\hline
Soccer Ball      & 0.858 & 0.900 & 0.946 & 0.716 \\
Goal Post        & 0.821 & 0.845 & 0.911 & 0.684 \\
Robot            & 0.349 & 0.368 & 0.247 & 0.123 \\
L-Intersection   & 0.664 & 0.621 & 0.529 & 0.115 \\
T-Intersection   & 0.764 & 0.719 & 0.712 & 0.198 \\
X-Intersection   & 0.682 & 0.586 & 0.543 & 0.150 \\
\hline
\textbf{Overall} & 0.690 & 0.673 & 0.648 & 0.331 \\
\hline
\end{tabular}
\end{table}

As shown in \cref{tab:vision_results}, the model is intentionally tuned toward high precision to avoid spurious detections, which are particularly harmful for localization. This leads to conservative recall, especially for field landmarks, but ensures that detected landmarks are highly reliable. The soccer ball class received additional targeted tuning, resulting in strong performance across all metrics to support robust gameplay-critical perception.

\subsubsection{Proximity}

Proximity sensing performance was evaluated using precision, recall, F1 score, and distance error. To reduce wall-induced false positives, detections within 0.5\,m of field boundaries (goal line at $x=7$\,m, side boundaries at $y=\pm4$\,m, and start line at $x=0$\,m) were filtered out, though occasional detections of the human handler remained. Each remaining detection was matched to the nearest ground-truth obstacle by computing the Euclidean distance to the closest point on its $0.4\,\text{m} \times 0.4\,\text{m}$ bounding box. A match within 0.2\,m was counted as a true detection; larger deviations were treated as false positives, and missed obstacles as false negatives.

\cref{fig:subsystem result} A illustrates a representative run, where the robot traverses a manual path while detecting two static obstacles. The detected obstacle points cluster tightly around the ground-truth regions, and the localization trajectory shows that the robot consistently re-detects both obstacles throughout the motion. This qualitative behavior aligns with the quantitative results in \cref{tab:proximity_results}.

F1 score was computed as the harmonic mean of precision and recall, and spatial accuracy was quantified using the mean and standard deviation of distance errors across true detections. As summarized in \cref{tab:proximity_results}, the system achieves high precision, perfect recall, and low spatial error (mean $1.9$\,cm), demonstrating reliable and accurate proximity sensing under realistic on-field conditions.

\begin{table}[htbp]
\centering
\caption{Proximity results across 3 runs. Metrics include Precision (P), Recall (R), F1 Score (F1), Distance Error (E).}
\label{tab:proximity_results}
\begin{tabular}{lcccc}
\hline
\textbf{Run} & \textbf{P} & \textbf{R} & \textbf{F1} & \textbf{E}\\
\hline
1        & 0.971 & 1.000 & 0.9850 & $0.011 \pm 0.028$ \\
2        & 0.9408 & 1.000 & 0.9695 & $0.024 \pm 0.048$ \\
3        & 0.911 & 1.000 & 0.9534 & $0.022 \pm 0.040$ \\
\hline
\textbf{Avg} & $0.94\pm0.02$ & $1\pm0$ & $0.97\pm0.01$ & $0.019\pm0.01$ \\
\hline
\end{tabular}
\end{table}

\subsubsection{Localization}
Reliable localization is essential for decision making, path planning, and kick execution in dynamic RoboCup environments. During the competition, both the localization node and the complementary particle-filter node ran at a stable 100~Hz in ROS~2, even under multiple detections and $K$-means clustering. To mitigate latency spikes, the YOLO-v8 perception module was limited to a maximum of seven detected landmarks per frame. The vision subsystem exhibited an approximately linear relationship between distance and percentage error in landmark range estimation: $\pm 1\%$ for distances $<2\,\mathrm{m}$, $\pm 3\%$ for $<4\,\mathrm{m}$, $\pm 5\%$ for $<6\,\mathrm{m}$, $\pm 8\%$ for $<8\,\mathrm{m}$, and $\pm 10\%$ for distances $>8\,\mathrm{m}$. Ground-truth robot pose was obtained using a Vicon motion-capture system operating at 300~Hz.

\cref{fig:subsystem result} B visualizes the isolated localization test, where CLAP estimates (\textit{purple}) are compared against Vicon ground truth (\textit{orange}) as the robot repeatedly traces a five-pointed star trajectory. For clarity, a single lap is shown. Due to motion-capture volume limits, the evaluation was performed in the square region on the right side of the field; however, this encompasses sufficient variation in heading and position to test robustness. Across the full trajectory, CLAP achieves a mean squared error (MSE) of $0.0357~\mathrm{m}^2$, a mean absolute error (MAE) of $0.1651~\mathrm{m}$, and a standard deviation of $0.0916~\mathrm{m}$. These results demonstrate that the geometric CLAP method provides stable, drift-resistant localization under realistic conditions, maintaining accuracy despite landmark sparsity, detection noise, and field symmetry.

\subsubsection{Navigation}
We evaluated the navigation stack in two stages: pure trajectory tracking and integrated navigation with obstacle avoidance.

First, to isolate tracking performance, the robot was commanded to follow a predefined five-pointed star trajectory without obstacles, as shown in \cref{fig:subsystem result} C. The reference path is drawn in \textit{blue}, and the executed path from motion capture is shown in \textit{orange}. the tracking error statistics are $\mathrm{MSE}=0.0141~\mathrm{m}^2$, $\mathrm{MAE}=0.0945~\mathrm{m}$, and $\mathrm{STD}=0.0722~\mathrm{m}$, indicating sub-decimeter average deviation and highly consistent cf-MPC tracking across the entire  pattern. Panels B and C share the same nominal trajectory; although the localization trace in \cref{fig:subsystem result} B appears noisier, the executed path in \cref{fig:subsystem result} C is visibly smoother. This is expected, as the combination of cf-MPC and the robot’s physical dynamics effectively acts as a low-pass filter on noisy state estimates, attenuating high-frequency localization jitter while still following the underlying path.
 
We next enabled full navigation (planning + tracking) in the presence of obstacles, as shown in \cref{fig:subsystem result} D. Two virtual circular obstacles of radius 1.25\,m were added, and the robot was commanded to walk from the origin toward a goal region on the right side of the field. Due to the limited motion-capture volume (red dashed box), only the central portion of each trajectory was recorded, which still fully covers the obstacle-avoidance region. Six independent trials (colored curves) were conducted. Across all runs, the robot generated smooth, dynamically feasible paths that bend around both obstacles while maintaining boundary clearance. The mean path (orange) lies well within the shaded $\pm 1\sigma$ band, indicating low trial-to-trial variation and strong repeatability of the navigation pipeline. At worst, trajectories barely touch the circular keep-out boundaries, reflecting modest tracking and estimation error while remaining within the configured safety margin.

Together, these tests demonstrate that our navigation method provides accurate trajectory tracking and robust, real-time obstacle-aware navigation on a full-scale humanoid platform.

\subsubsection{Kick}
To assess the accuracy and consistency of the proposed PLMK kick, we placed balls at multiple locations across the field and commanded ARTEMIS to autonomously shoot toward the center of the goal. For each test condition we executed 10 kicks, and repeated this over 5 groups (50 kicks total). Ball trajectories were recorded using an overhead vision system, from which we fitted the initial kick direction and recovered the full ball path.

For graphical clarity, \cref{fig:subsystem result} E, F show only one representative group of 10 kicks; the other groups exhibit similar behavior. In \cref{fig:subsystem result} E, the fitted initial directions cluster tightly inside the preset $\pm 15^\circ$ kick cone. Across all 50 kicks, 46 remained within this cone, indicating that the perception-locked strike point and mid-swing foot retargeting reliably constrain launch direction despite variations in gait phase and perception noise. \cref{fig:subsystem result} F overlays the corresponding ball paths on the field: all kicks are aimed at the geometric center of the goal, and 42 of 50 result in successful goals. Taken together, these results show that the PLMK module delivers accurate, repeatable, and directionally constrained in-gait kicks suitable for real-time soccer play.

The PLMK also highlights the dynamic advantage of ARTEMIS’s hardware and gait design. With a swing duration of 0.4~s, the leg reaches peak linear velocity of approximately 3.2~m/s while kicking a ball. \cref{tab:kick_performance} compares the kicking characteristics of ARTEMIS with representative humanoids using other kick methods\cite{2022nimbro}. Our method achieves high-speed, in-gait kicks without interrupting locomotion, leveraging low distal leg inertia and torque-dense proprioceptive actuators.

\begin{table}[t]
  \caption{Comparison of kicking dynamics across humanoid platforms}
  \label{tab:kick_performance}
  \centering
  \begin{threeparttable}

  \setlength{\tabcolsep}{6pt}
  \renewcommand{\arraystretch}{1.1}

  \begin{tabular}{@{}l l c@{}}
    \toprule
    \textbf{Robot} & \textbf{Kick Type}
      & \textbf{Impact Foot Speed(m\,s$^{-1}$)} \\
    \midrule
    ALICE         & Static ZMP          & 0.8\tnote{*} \\
    NimbRo-OP2X   & Semi-Swing          & 1.2 \\
    \textbf{ARTEMIS (ours)}& \textbf{PLMK} & \textbf{3.2} \\
    \bottomrule
  \end{tabular}

  \begin{tablenotes}[flushleft]
    \item[*]\footnotesize Data based on video estimation.
  \end{tablenotes}
  \end{threeparttable}
\end{table}

\subsection{Foot Attachment Material Simulation}
\begin{figure}[htbp!]
    \centering
    \includegraphics[width=\linewidth]{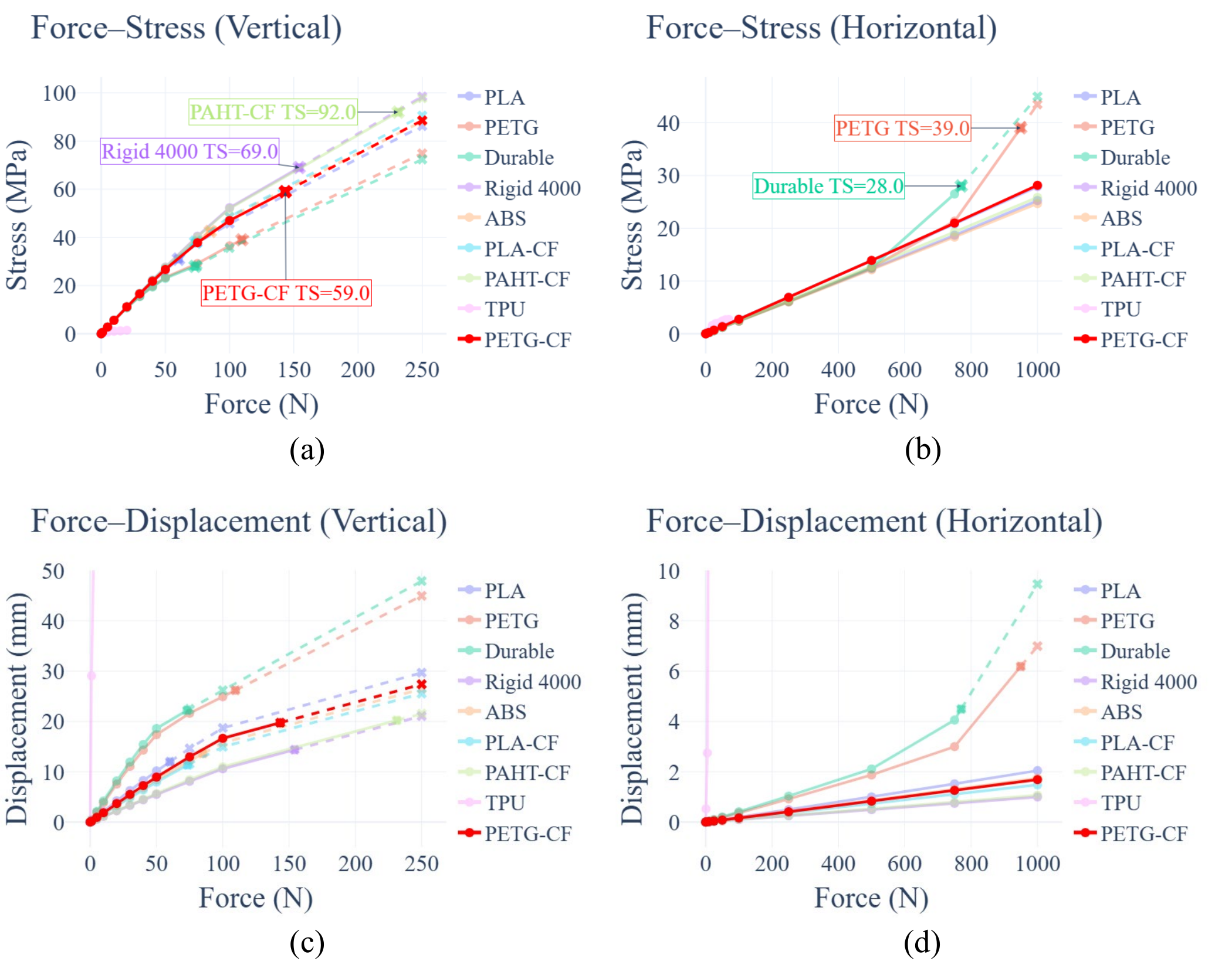}
    \caption{FEA results for the foot module: (a) Vertical Force-Stress, (b) Horizontal Force-Stress, (c) Vertical Force-Displacement (TPU simulation failed when displacement reached 125.74mm), and (d) Horizontal Force-Displacement (TPU simulation failed when displacement reached 98.017mm).}
    \label{fig:foot_fea}
\end{figure}

For the foot design, we conducted material testing to evaluate stiffness and directional performance. Since the key functional requirement of the foot attachment is to maintain high strength in the horizontal direction while preserving flexibility in the vertical direction, we focused on analyzing the relationship between applied force, deformation, and resulting stress. A total of nine different materials were tested using Finite Element Analysis (FEA) in ANSYS to simulate their behavior under directional loading conditions. The results of this comparative study are presented in.\cref{fig:foot_fea}.

Among all the materials tested, PETG-CF was ultimately selected for the foot attachment due to its balance between mechanical performance and ease of manufacturing. As shown in \cref{fig:foot_fea} (a) and (b), PETG-CF exhibits excellent durability under horizontal loading, withstanding forces up to 1000 N without failure, while fracturing under vertical loads exceeding approximately 150 N. The simulation range for horizontal forces was intentionally extended to reflect potential collisions with other robots during real matches, where the foot attachment may impact rigid components such as plastic or metal covers. Vertically, PETG-CF demonstrated higher impact tolerance compared to other materials, meeting the design objective under excessive external impact. Among the alternative materials evaluated, TPU was too flexible to provide controllable fracture under vertical load which further impact the locomotion (\cref{fig:foot_fea} (c) and (d)), while Rigid 4000 required longer printing and post-processing time, and PAHT-CF is less cost-effective and more sensitive to the printer setup. These crucial shortcomings make PETG-CF the most balanced choice in terms of mechanical performance, manufacturability, and operational practicality.






\subsection{High-level Strategy Simulation}

\begin{figure}[htbp]
    \centering
    \includegraphics[width=0.48\textwidth]{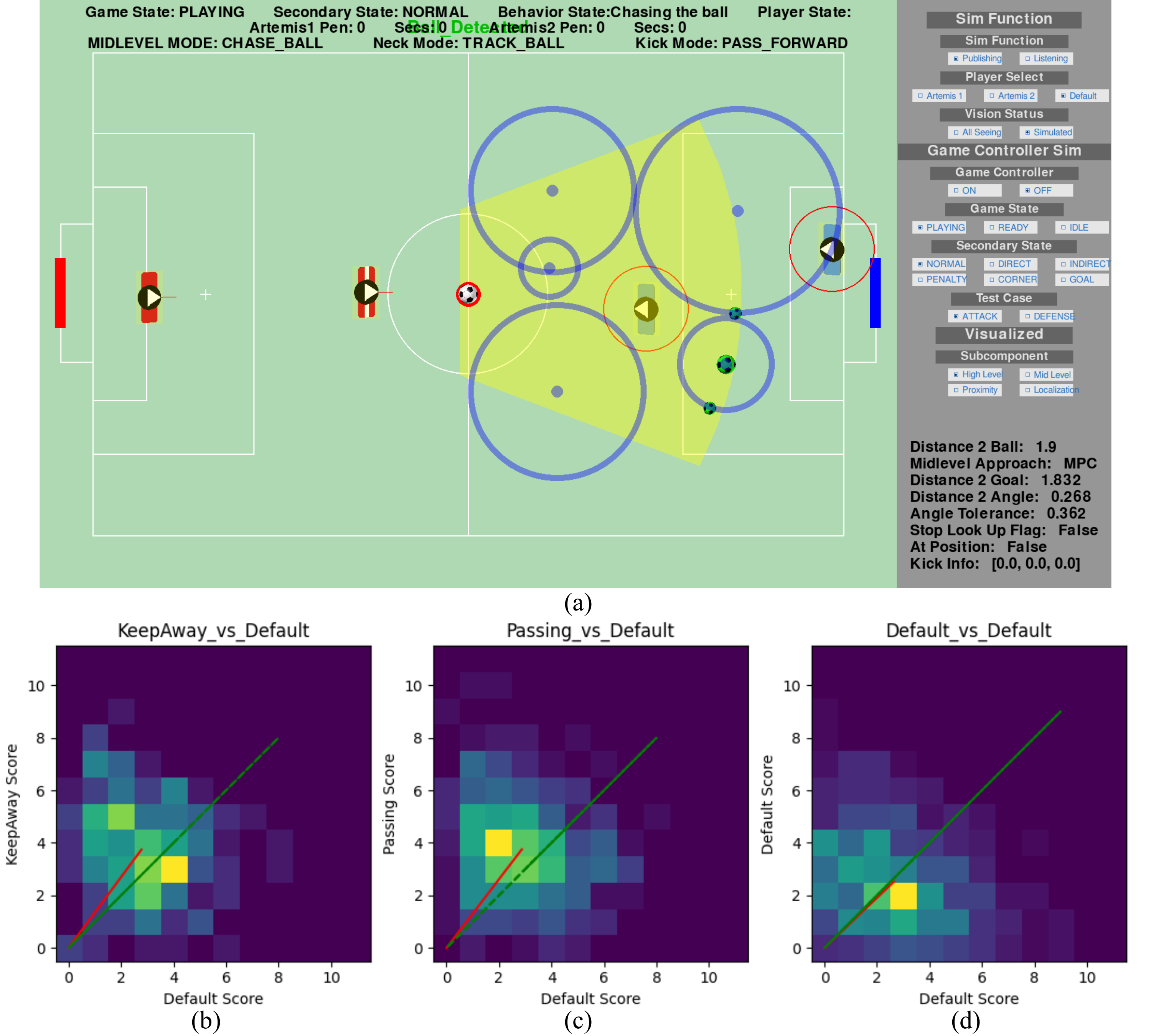}
    \caption{Simulation Environment and results of high level strategies. a). A snapshot of simulated game visualizer. b) Kick away from opponents, c). Kick closest to goal, d). Shoot on goal. The histograms indicate the relative number of instances of a particular score outcome of the simulated matches. }
    \label{fig:simulated}
\end{figure}

\subsubsection{Simulation Setup}

When developing our high level logic, debugging on physical hardware was challenging due to sensor noise, environmental variability, and errors from other components. This motivated the creation of custom software debugging tools. Initially we created a Field Logic GUI (\cref{fig:simulated} (a)), implemented as a ROS node that mimicked the outputs of the Vision and Localization modules, allowing us to test behavior tree decisions under controlled scenarios. While effective for verifying basic functionality, this tool offered limited insight into how to improve overall system behavior. 

To accomplish this we created a simple Game Simulator alongside the GUI. Within the simulator, the position and velocity of the ball and each player was tracked. The navigation module was utilized to update the player states at each timestep and a simple dynamics model was applied to the ball. This model also included a consideration of the interaction of the ball and players, particularly kicks and unintentional collisions. The rules of the game were coded the same as the real competitions. We were able to replicate the signals of a full game in real time with this simulator. From here we could better evaluate the effectiveness of different strategies to begin to optimize the module with quantitative comparisons.

\subsubsection{Optimization}
We began the analysis by examining our passing algorithm, which determines the best target location based on a geometric assessment of the field. To study its behavior, we created several parameter profiles that encouraged different strategies, such as prioritizing shots on goal or selecting open passing spaces, each producing distinct gameplay styles. However, variability in robot motion and ball dynamics introduces significant uncertainty, meaning that a large dataset is required to accurately characterize the distribution of likely outcomes.

During the simulation, we tested two proposed strategies against the default approach of always shooting on goal. As shown in \cref{fig:simulated} (b), kicking the ball away from opponents proved more effective overall than shooting directly, while its most frequent outcome was a loss (the yellow region). On average, however, it achieved better performance, as indicated by the higher mean outcome (the red line). For \cref{fig:simulated} (c), passing strategy exhibited a stronger outcome than which of the default strategy, often resulting in a 4–3 victory, and with a better average performance. The result in \cref{fig:simulated} (d) served as a baseline to verify the symmetry and correctness of the simulation setup. These simulations provide qualitative insight into which strategies are most advantageous under different match conditions.

\subsection{Full-Stack Autonomous Soccer Performance}

\begin{figure*}[htbp]
    \centering
    \includegraphics[width=0.8\linewidth]{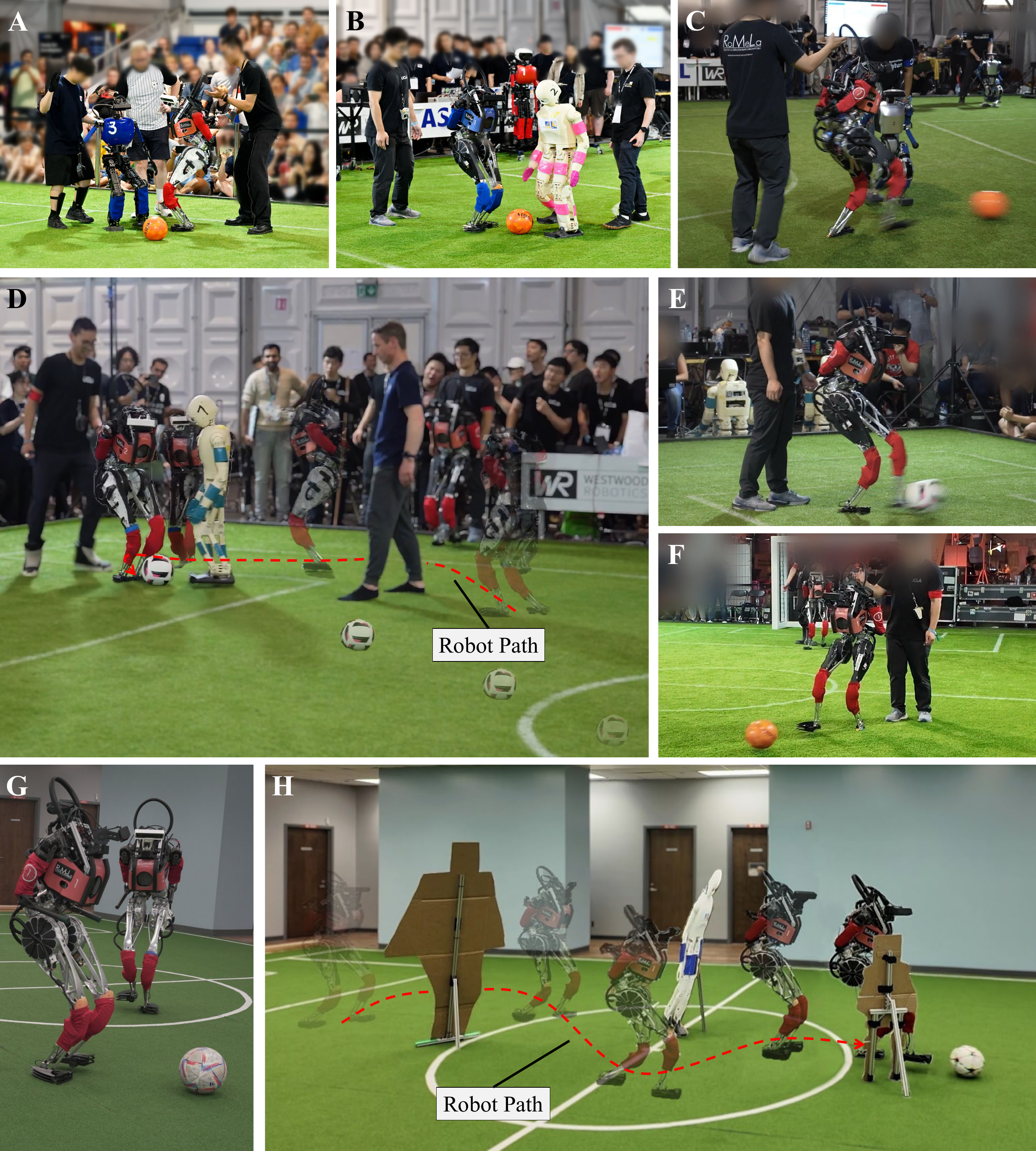}
    \caption{Full stack soccer performance. A)-C). Closed range interaction with opponent robots. D). Navigation bypassing opponents and kick the ball. E) F). Strong kick within short range and long range. G). Kick-off team cooperation. H). Pseudo-game with random opponents and ball positions for system performance test.}
    \label{fig:real_test}
\end{figure*}

The system performance is ultimately verified during the RoboCup 2024 competition. As the champion of the division, our team scored a total of 45 goals over 6 official matches. Under competition rules, all robots must operate fully autonomously, maintain human-like physical proportions, and rely only on sensors with biological analogs, such as cameras as visual sensors while devices such as magnetometers are prohibited. In addition, discrepancies between practice fields and the official competition field introduce further challenges for reliable gameplay. Despite these constraints, the proposed system enabled ARTEMIS to demonstrate stable locomotion across different terrains, robust perception and localization under varying illumination conditions and field sizes, effective obstacle avoidance in highly dynamic soccer environments, and reliable passing and scoring performance throughout the matches.

\cref{fig:real_test} illustrates representative scenarios from real match evaluation. \cref{fig:real_test} A-C illustrate close range interactions, where the system enabled the robot to avoid collisions with opponents, ensure the safety of nearby humans, and make correct decisions for interacting with the ball. In \cref{fig:real_test} D, our robot efficiently bypassed the opponents, and reached a suitable position for clearance. The kick performance is demonstrated in \cref{fig:real_test} E and F. Benefiting from the hardware design and high level kick logic, the robot successfully scored both from close range near the goal, and positions around the field center its own side. \cref{fig:real_test} G illustrates the kickoff cooperation scenario, where one robot performed a slight initial tap of the ball while another accelerated forward with maximum speed to execute the goalscoring kick. \cref{fig:real_test} H presents a pseudo-game scenario designed to evaluate the overall system performance. In this test, opponent robots were placed at random positions, and the ball was positioned within the robot’s field of view. The robot successfully detected the ball, navigated through the opponents without collision, reached the desired position, and eventually scored a goal.

These evaluations demonstrate the reliability of the proposed system under highly dynamic and adversarial competition conditions with varying field environments. The robot consistently exhibited stable locomotion, accurate perception and localization, and robust collision-free navigation. A wide range of behaviors were successfully executed during real matches, and these results collectively show that the tight integration of hardware and software enables ARTEMIS to maintain high performance, resilience, and strategic effectiveness in real competition settings.

\section{Discussion} 
\label{sec:dis}

The results demonstrate that the tightly coupled, model-based system plays a central role in the robust performance of ARTEMIS during RoboCup competitions. Unlike loosely connected modular pipelines, the direct interaction among perception, localization, behavior tree, path planning, and locomotion enables consistent information flow and fast closed-loop responses to dynamic game situations. Perception provides foundational situational awareness, which is immediately leveraged by localization for pose estimation, by the behavior tree for decision making, and by the planner for collision-aware trajectory generation. This tight coupling allows velocity and kicking commands to be executed reliably at the hardware level, resulting in dynamically consistent locomotion and effective task execution under frequent disturbances and opponent interactions. Moreover, the hardware design is optimized with the software stack, particularly in the structural components and foot attachment, ensuring that the robot can physically realize the actions commanded by the higher level controllers.

Nevertheless, several limitations still remain. The current behavior layer relies on deterministic rules, which constrain adaptability in rapidly changing scenarios. In addition, the performance of the upper layers is ultimately bounded by perception and localization robustness, particularly under severe occlusions and cluttered conditions. These observations suggest that future improvements should focus on more adaptive decision making strategies, stronger multi-sensor fusion for localization, and tighter feedback between high level planning and low level control to further enhance agility, robustness, and autonomy in dense competitive environments.

Recent work from Tsinghua University \cite{booster_2025learning} shows that learning-based methods can generate highly reactive, vision-driven soccer behaviors, exceeding traditional model-based approaches in adaptivity and simplicity. However, such methods remain sensitive to sim-to-real gaps, offer limited interpretability, and require substantial data and computation. In contrast, model-based methods provide stable, predictable performance and easier debugging but are constrained by modeling assumptions. Regarding these complementary strengths and weaknesses, hybrid approaches that preserve model-based robustness while incorporating learning-driven adaptability represent a promising direction for future development, where we also aim to pursue.

\section{Conclusion and Future Work}

In this paper, we presented an integrated model-based system for humanoid soccer and evaluated its performance through simulation, field tests, and official RoboCup competition. We described our novel foot attachment design and the complete perception, localization, navigation, and behavior framework. The fully integrated system demonstrated reliable localization, efficient path planning and obstacle avoidance, and effective cooperative passing and goal-scoring strategies, validating the robustness and competitiveness of ARTEMIS in real match environments.

Future work will focus on improving both hardware robustness and software intelligence. On the hardware side, we aim to develop an active upper body for autonomous fall recovery and improved protection, explore multi-material fabrication for more durable and compliant foot attachments that is adaptable to different field conditions, and upgrade the battery and power distribution system for higher peak current handling and longer runtime. On the software side, we aim to extend current model-based stack and combine learning-based decision-making, localization, and path planning, integrate predictive whole body control for improved agility and balance, and advance multi robot coordination toward cooperative team behaviors.

\bibliographystyle{IEEEtran}
\bibliography{references}
\end{document}